\documentclass[10pt,twocolumn,letterpaper]{article}

\usepackage{wacv}              

\usepackage{graphicx}
\usepackage{amsmath}
\usepackage{amssymb}
\usepackage{booktabs}

\usepackage[pagebackref,breaklinks,colorlinks]{hyperref}

\usepackage[capitalize]{cleveref}
\crefname{section}{Sec.}{Secs.}
\Crefname{section}{Section}{Sections}
\Crefname{table}{Table}{Tables}
\crefname{table}{Tab.}{Tabs.}

\usepackage{algorithm}  
\usepackage{algpseudocode}  
\usepackage{caption}
\usepackage{subcaption}
\usepackage{pifont}
\usepackage{makecell}
\usepackage{colortbl}
\usepackage[table]{xcolor}
\usepackage{stfloats}
\usepackage{dirtree}
\usepackage[normalem]{ulem}
\usepackage[accsupp]{axessibility}  
\apptocmd{\dirtree}{\bigskip}{}{}
\pretocmd{\dirtree}{\bigskip}{}{}

\def \ninstances {$2,370$}
\def \nyieldinstances {$351$}
\def \nplots {$388$}
\def \nyieldplots {$145$}
\def \nimages {$209,000$}
\def \npaddy {$1,007$}
\def \ncroptype {$21$}
\def \dataset {\texttt{\textbf{\textit{SICKLE}}}}
\def \rcolor {green!20}
\newcommand{\cmark}{\ding{51}}%
\newcommand{\xmark}{\ding{55}}%

\def\wacvPaperID{1083} 
\def\confName{WACV}
\def\confYear{2024}

\begin{document}

\title{SICKLE: A Multi-Sensor Satellite Imagery Dataset \\Annotated with Multiple Key Cropping Parameters}

\author{Depanshu Sani$^1$, Sandeep Mahato$^2$, Sourabh Saini$^1$, Harsh Kumar Agarwal$^1$, \\
Charu Chandra Devshali$^2$, Saket Anand$^1$, Gaurav Arora$^1$, Thiagarajan Jayaraman$^2$ \\
$^1$Indraprastha Institute of Information Technology, Delhi, India\\
$^2$MS Swaminathan Research Foundation, Chennai, India\\
{\tt\small \url{https://sites.google.com/iiitd.ac.in/sickle/home}}
} 
\maketitle

\begin{abstract}
   The availability of well-curated datasets has driven the success of Machine Learning (ML) models. Despite greater access to earth observation data in agriculture, there is a scarcity of curated and labelled datasets, which limits the potential of its use in training ML models for remote sensing (RS) in agriculture. 
   To this end, we introduce a first-of-its-kind dataset called \dataset{}, which constitutes a time-series of multi-resolution imagery from $3$ distinct satellites: Landsat-$8$, Sentinel-$1$ and Sentinel-$2$. Our dataset constitutes multi-spectral, thermal and microwave sensors during $\text{January }2018-\text{March }2021$ period. We construct each temporal sequence by considering the cropping practices followed by farmers primarily engaged in paddy cultivation in the Cauvery Delta region of Tamil Nadu, India; and annotate the corresponding imagery with key cropping parameters at multiple resolutions (i.e. $3$m, $10$m and $30$m).
   Our dataset comprises \ninstances{} season-wise samples from \nplots{} unique plots, having an average size of $0.38$ acres, for classifying \ncroptype{} crop types across $4$ districts in the Delta, which amounts to approximately \nimages{} satellite images. Out of the \ninstances{} samples, \nyieldinstances{} paddy samples from \nyieldplots{} plots are annotated with multiple crop parameters; such as the variety of paddy, its growing season and productivity in terms of per-acre yields.
   Ours is also one among the first studies that consider the growing season activities pertinent to crop phenology (spans sowing, transplanting and harvesting dates) as parameters of interest. 
   We benchmark \dataset{} on three tasks: crop type, crop phenology (sowing, transplanting, harvesting), and yield prediction. 
\end{abstract}
\section{Introduction}
\label{sec:intro}

\begin{figure}
    \centering
    \includegraphics[trim={2cm, 1cm, 6cm, 0}, clip, width=0.74\linewidth]{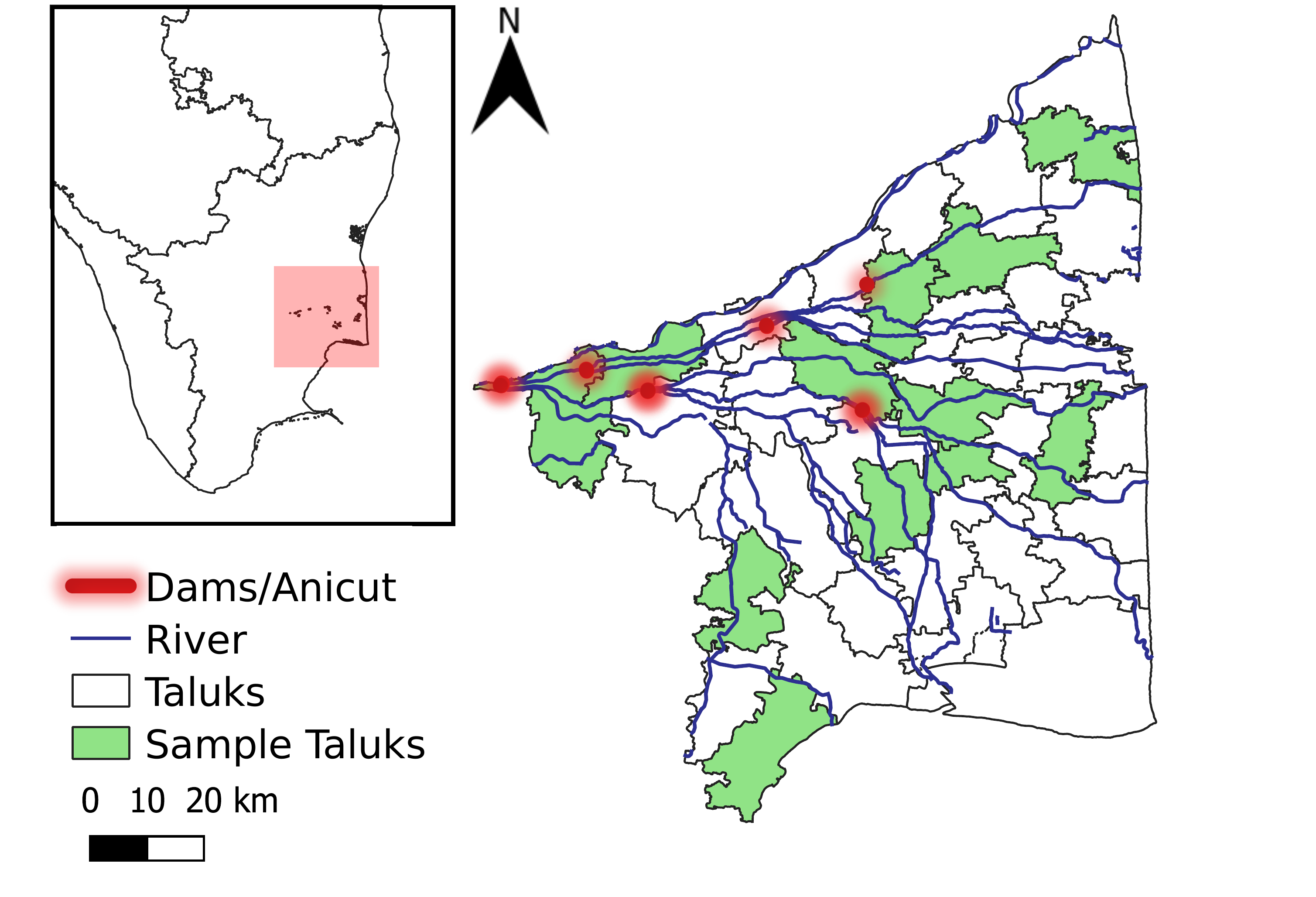}
    \caption{Schematic diagram of the complete study region in Tamil Nadu, India. Field data was collected from the highlighted blocks.}
    \label{fig:intro-diagram}
\end{figure}

Satellite imagery has emerged as a widely applicable RS tool for researchers across domains like wildlife conservation, climate science, and economics among others; due to its ability to generate policy-relevant data for large and inaccessible areas in a cost-effective manner. Despite the abundance of satellite data, there is a lack of curated datasets that are annotated with field data, thereby limiting its potential applicability for training and evaluating ML models. Agriculture is also one such domain where collecting field data involves conducting ground-based surveys, which is expensive and laborious. There exist publicly available datasets that target specific tasks, such as crop identification \cite{cropharvest, sustainbench, radiant}, land cover classification \cite{radiant, sustainbench} and yield estimation \cite{sustainbench} among others. These datasets provide value when researchers are interested in singular line of inquiry that requires one dataset at a time. For instance, a crop yield prediction model can be trained using \textsc{SustainBench}'s crop yield dataset \cite{sustainbench} and a transfer learning approach can then be adopted. However, the dataset only has county-level annotations, which precludes a high-resolution, plot-level analysis.
Furthermore, most available datasets have different characteristics and cover different geographies; for instance, the PASTIS \cite{garnot2021panoptic} has multi-spectral information for crop type mapping within the French metropolitan territory; whereas Agriculture-Vision \cite{agriculture-vision} has only Red, Green, Blue and NIR spectral bands for crop anomaly detection across the US; and \textsc{SustainBench}'s field delineation task has only visible spectrum bands in France. Therefore, it is not possible to train and evaluate a single model for different, yet closely related tasks, e.g., harvesting dates and yield estimates. Moreover, the time-series samples in these datasets are produced using an arbitrary sequence length, which might affect the overall performance. 
For example, each sequence in PASTIS-R \cite{pastis-r} contains observations taken between $\text{Sep'}18-\text{Nov'}19$ and has no details regarding the crop phenology dates. It is likely that multiple crops have been cultivated in this period and hence the quality of input is degraded, ultimately affecting the performance. All the aforementioned datasets lack labels of key cropping parameters, which, if predicted accurately, could lead to better-informed downstream decisions, whereas \cite{kehs2021busia} provides the key cropping parameters but lacks the access to satellite imagery data, which adds an overhead to collect and process the data from multiple satellites making the training process cumbersome.
To the best of our knowledge, there exists no prior dataset that contains multi-sensor satellite data annotated with multiple cropping parameters for the same set of plots.
To tackle the mentioned challenges, we make the following contributions that can significantly impact the quality of deployment and hence promote applicability of RS and ML in agriculture:
\begin{enumerate}
    \item  We introduce a first-of-its-kind dataset, \dataset{} (\textbf{S}atellite \textbf{I}magery for \textbf{C}ropping annotated with \textbf{K}ey-parameter \textbf{L}ab\textbf{E}ls), containing images from multiple satellites (Landsat-$8$, Sentinel-$1$ and Sentinel-$2$) having a variety of sensors (optical, thermal and radar) with annotations of multiple agricultural cropping parameters for each plot. These annotations are created at $3$ scales, i.e., $30m$, $10m$ and $3m$ spatial resolution.
    \item We organize the dataset in a way that can be readily used by researchers from multiple domains, including agronomy, RS and ML. Multiple cropping parameters for the same set of plots allow multi-task learning, which can help learn better feature representation. The availability of multi-scale annotations enables the generation of high-resolution (HR) inference maps from low-resolution (LR) images. Moreover, the individual bands at actual resolutions and the metadata are provided, which can help develop a better band interpolation strategy. The presence of time-series image sequences from different satellites of the same geographical location can help evolve cross-satellite fusion, synthetic band generation and forecasting techniques.
    \item We include the plot-level crop phenology dates, i.e., sowing, transplanting and harvesting dates, in the dataset. In the absence of these dates, the estimation of cropping parameters using time-series data can suffer due to interference from other seasonal crops planted in cases of heterogeneous farming. 
    \item We present a novel strategy for preparing time-series data over a seasonal temporal window that is consistent with the \textit{regional cropping standards} that are typically followed by the farmers  for crop production, which improves the robustness and correctness of a solution in a real-world deployment.
\end{enumerate}
\section{Related Work}
\label{sec:lit-survey}

\subsection{Crop Type Mapping}
With the increased application of multi-temporal datasets for crop type mapping, the classification methods have transitioned from applying Maximum Likelihood based approach \cite{yu2014} to modern algorithms such as Support Vector Machines \cite{mathur2008,kumar2015,virnodkar2020}, Decision Trees \cite{deschamps2012,wei2018}, Random Forest \cite{hariharan2018,long2013object} and neural networks \cite{zhao2019,castro2020rice}. 
However, the availability of such datasets is limited. Often, there is an imbalance with respect to the availability of labelled datasets across geographical regions. For instance, Europe has an abundance of large, densely annotated datasets, whereas, for regions like South Asia and Africa, labelled datasets are sparsely available.
Kehs et al. \cite{kehs2021busia} provide a field dataset collected from $\text{May } 6, 2019-\text{June } 9, 2019$ having plot boundaries, crop type labels, irrigation status, density of green leaf area and other key parameters for $474$ plots in Northern Busia county, Kenya. CV4A Kenya Crop Type Competition \cite{REF} dataset with segmented sentinel-2 tiles is also available for Kenya. For Ghana and Sudan, \textsc{SustainBench} provides data inputs as growing season time series of imagery from three satellites: Sentinel-$1$, Sentinel-$2$, and PlanetScope in $2016$ and $2017$ for different crop types \cite{sustainbench}. Some other datasets published particularly focusing on crop type target class for the European regions are \cite{russwurm2019breizhcrops,pastis-r,sykas2022sentinel}. At the global level CropHarvest collection is available with crop/non-crop and agricultural class labels, collected from publicly available datasets and covering $343$ labels \cite{cropharvest}. Except for \cite{kehs2021busia} all the datasets listed here are labelled with crop type information only, but on the other hand, \cite{kehs2021busia} has no satellite imagery associated with it. 

\subsection{Crop Phenology Dates}
Estimating start and end of the season requires phenology metrics to be derived from satellite images. 
Methods such as shape model fitting of time-series indexes such as NDVI and EVI are widely used for generating phenology metrics of a crop season \cite{singha2016object,liu2022detecting}. For paddy crops, SAR-based time-series are also applied to derive these season dates \cite{wang2022parcel}. Phenology dates are often used for rule-based crop type classification \cite{singha2016object}.
The training data provided by \cite{kehs2021busia} has labels on the planting and harvest dates for Maize at plot level. However, this data has not been benchmarked yet, possibly due to unavailability of any satellite images with the dataset. 

\subsection{Crop Yield}
Current methods for yield estimation use empirical as well as process-based models. ML-based empirical models can produce reliable estimates of paddy yield \cite{palanivel2019approach}. Random Forest models with Sentinel-$1$ data have been used to predict paddy yield with high accuracy \cite{clauss2018estimating}. \cite{sharma2020wheat} used CNN-LSTM with Sentinel-$2$ images for tehsil (block) level wheat predictions and achieved $50\%$ better performance over other models. The Process-based models mentioned here use dynamic crop models with remote sensing inputs along with other parameters to simulate crop yield.
\textsc{SustainBench} crop yield datasets provides county-level yields for $857$ counties in the United States, $135$ in Argentina and $32$ in Brazil for the years $2005-2016$ \cite{wang2018transfer}. 


\subsection{Lack of Multiple Downstream Tasks}
There is a lack of publicly available datasets that can be used for predicting multiple key cropping parameters. 
One of the largest datasets available for multiple tasks, such as crop type mapping and field delineation, is \cite{sustainbench}. However, the geographical coverage of these datasets varies depending on the specific tasks involved. In contrast, both semantic and panoptic segmentation masks are supplied by \cite{garnot2021panoptic, pastis-r} for a specific area; a feature that is lacking in \cite{sustainbench}. Addressing anomaly patterns, plot boundaries along with nine distinct types of these patterns are provided by \cite{agriculture-vision}.
Despite the fact that few available datasets provide multiple cropping patterns, neither of them can be used to train a multi-tasking end-to-end trainable network for reasons outlined above.
\section{Dataset Description}
The Cauvery Delta is a major rice cultivation region in Tamil Nadu, India, where farmers cultivate one or two crops of paddy, depending upon water availability. 
However, due to significant shifts in the its agrarian landscape, an Indian Council for Agricultural Research (ICAR) study in $2013$ reclassified four districts of the Cauvery Delta from dry semi-humid to semi-arid conditions \cite{ICAR_report}\footnote{The supplementary material has more details about the study region, data sampling and dataset statistics.}.
Given its centrality to rice production in India, the inferences via well-designed ML methods from this dataset would enable analysis of the impact of aridification on paddy yields and other cropping patterns in the region. Moreover, we anticipate that the challenges of the \dataset{} dataset (discussed hereafter in section \ref{sec:challenge}) will also drive novel vision techniques.
We highlight the key features of \dataset{}, mentioned in section \ref{sec:intro}, and show a detailed comparison with the recent literature in  Tables \ref{tab:tasks} and \ref{tab:characteristics}. 

\begin{table*}
  \begin{center}
    {\small{
\begin{tabular}{lcccccccc}\toprule
\textbf{Tasks} & \textbf{\cite{sustainbench}} &\textbf{\cite{radiant}} &\textbf{\cite{agriculture-vision}} &\textbf{\cite{pixel-set}} &\textbf{\cite{pastis-r}} &\textbf{\cite{cropharvest}} &\dataset{} \\\midrule
Crop Type Semantic Segmentation & \cmark & \cmark &  & \cmark & \cmark & \cmark & \cmark \\
Crop Type Panoptic Segmentation &  &  &  &  & \cmark &  & \cmark \\
Cropland Segmentaion & \cmark & \cmark &  &  &  &  &  \\
Field Delineation & \cmark &  & \cmark &  &  &  & \cmark \\
Phenology Date Prediction &  &  &  &  &  &  & \cmark \\
Crop Yield Prediction & \cmark &  &  &  &  &  & \cmark \\
Crop Anomaly Detection &  &  & \cmark &  &  &  &  \\
\arrayrulecolor{black!30} \midrule 
All above tasks for same set of plots & \color{red}{\xmark} & \color{red}{\xmark} & \color{green}{\cmark} &  & \color{green}{\cmark} &  & \color{green}{\cmark} \\
\arrayrulecolor{black!30} \midrule 
Multi-Task Learning &  &  & \cmark &  &  &  & \cmark \\
Multi Image Super Resolution &  &  &  &  &  &  & \cmark \\
Cross-Satellite \& Cross-Sensor Fusion &  &  &  &  & \cmark &  & \cmark \\
Synthetic Band Generation &  &  &  &  & \cmark &  & \cmark \\
HR prediction using LR images &  &  &  &  &  &  & \cmark \\
\arrayrulecolor{black}\bottomrule
\end{tabular}
}}
\end{center}
\caption{A comparison of \dataset{} with related datasets (\textsc{SustainBench}\cite{sustainbench}, Radiant ML Hub \cite{radiant}, Agriculture-Vision \cite{agriculture-vision}, Pixel Set \cite{pixel-set}, PASTIS-R \cite{pastis-r} and Crop Harvest \cite{cropharvest}) based on the tasks that can be performed using them. The bottom $4$ tasks are not only related to the agricultural domain but are also applicable for remote sensing community.}
\label{tab:tasks}
\end{table*}

\begin{table*}
  \begin{center}
    {\small{
\begin{tabular}{lcccccccc}\toprule
\textbf{Characteristics} & \textbf{\cite{sustainbench}} &\textbf{\cite{radiant}} & \textbf{\cite{pastis-r}} & \textbf{\cite{agriculture-vision}} & \dataset{} \\
\midrule
Time series data & \cmark & \cmark & \cmark &  & \cmark \\
Multiple annotations for same plots &  &  &  & \cmark & \cmark \\
Annotations at multiple resolutions &  &  & &  & \cmark \\
Multi-sensor data for all tasks & & & \cmark &  & \cmark \\
Consistent with the regional cropping practises &  &  &  &  & \cmark \\
Number of time-series samples & $[1,966-10,332]$ & Variable & $2,433$ & NA & \ninstances{} \\
\arrayrulecolor{black}\bottomrule
\end{tabular}
}}
\end{center}
\caption{A comparison of \dataset{} with the related datasets that can be used for multiple tasks 
(\textsc{SustainBench}\cite{sustainbench}, Radiant ML Hub \cite{radiant}, PASTIS-R \cite{pastis-r} and Agriculture-Vision \cite{agriculture-vision}) 
based on their characteristics.}
\label{tab:characteristics}
\end{table*}


\subsection{Data Acquisition}
Ground-based surveys were conducted throughout the study region to collect field data for the time span of $\text{January }2018-\text{March }2021$ for \nplots{} individual plots. We interacted with the farmers of these plots to collect information on the type and variety of the crop grown in each agricultural season, its growing season duration and productivity in terms of yield. Paddy being the primary focus of this study, the phenology dates and the crop yield were only gathered for the seasons in which paddy was cultivated. Moreover, the surveys were conducted from June $2021$ - February $2022$ and hence the collected field data is entirely dependent on the ability of the farmer to recall the details of the crops grown in the past. 
As a result, out of \ninstances{} samples, we have phenology dates and crop yield estimates for \nyieldinstances{} paddy samples only. For the remaining $656$ paddy samples, the farmers could only recall rough estimates of the season duration but not the crop yields. We also collected the GPS coordinates of the centroid of the plot, the block and the district it lies in, the relative location to the Cauvery Delta, and area of the plot in acres. The specific coordinates of each plot are withheld for privacy reasons. It is to be noted that ground-based surveys are expensive, inefficient and difficult to scale. Therefore, we surveyed only a subset of all the plots from multiple blocks at random.

For the entire study region, we acquired the tile images from $3$ publicly available satellites, Landsat-$8$, Sentinel-$2$\footnotemark{} and Sentinel-$1$, using the products `\path{LANDSAT/LC08/C02/T1_L2}', `\path{COPERNICUS/S2_SR}' and `\path{COPERNICUS/S1_GRD}' from the Google Earth Engine platform \cite{earth_engine}. Unlike previous works, we downloaded all the available bands, including the Quality Assessment and the derived bands, from all these satellites at their original resolutions band-wise, along with metadata. These bands and metadata files can be used as a prior for various tasks. For instance, Sentinel-$2$ provides a derived Scene Classification (SCL) band, which classifies each pixel as bare soil, vegetation, water, cloud, etc., that can provide more information about the image.
Now, if the crop cover map suggests that the water pixel (based on the prior) is a paddy crop, we can conclude that the model is not accurate.

\footnotetext{The surface reflectance (SR) data for Sentinel-$2$ is not available before Dec'$18$. Moreover, the top-of-atmosphere (TOA) data is also archived and is not accessible via Google Earth Engine. We requested the TOA tiles from Copernicus and converted them to SR images using Sen2Cor \cite{sen2cor}.}

\subsection{Data Annotation}
\label{sec:ann_technique}
For the annotation task, we created vector files for each of the \nplots{} surveyed plots using QGIS \cite{QGIS} by manually drawing the polygon boundaries around the GPS coordinates of each plot by visualizing the high-resolution image-based geographical maps available with the product. Since the broader objective of \dataset{} is to be able to generate inferences for the entire Cauvery region rather than a plot-wise analysis, we are interested in a region-wise analysis where the individual plots are unkown. Therefore, we create a rectangular buffer area of $320m\times320m$ around each plot. As many of our surveyed plots were from the same local region, the buffer area for such plots was highly overlapping. Thus, to regularize the relative locations of all the plots within these extended boundaries, we translated the centroid of the polygon buffers randomly in $X$ and $Y$ directions. These polygon buffer vectors were used for clipping out image patches from the satellite tile images. 
We identify all the plots within a buffer using the actual polygon vectors and then annotate all of them with \emph{plot\_id, crop\_type, sowing\_date, transplanting\_date, harvesting\_date and crop\_yield} labels at $30m$, $10m$ and $3m$ resolutions based on the preparation strategy mentioned in section \ref{sec:datasetpreparation}. For crop yield, we assume that cropping distribution was uniform within a plot and therefore distribute the amount of yield such that the sum of pixel-wise yield in a plot is equal to the total yield provided by the farmer.

\subsection{Dataset Preparation}\label{sec:datasetpreparation}
\textbf{Regional Standard Growing Season:} As discussed in section \ref{sec:intro}, using time-series data of an arbitrary sequence length will likely deteriorate the quality of cropping parameter prediction due to interference from crops planted in other seasons. If the time-series data spans an arbitrary duration ($t_a$) that is substantially smaller than the actual growing season of paddy ($t_p$), then the data might not be able to capture the complete phenological structure and hence will lack the information necessary for accurate prediction. On the other hand, if $t_a \gg t_p$, then the data will likely span multiple seasons and may witness interference due to a multi-crop cultivation leading to poor predictions. 
Therefore, we argue that the duration of the observations to create time-series input data should depend on the downstream task. Moreover, the crop phenology dates gathered from the farmers might differ from that region's standard paddy season duration. For instance, while the standard growing season starts in September, a farmer might sow the seeds in August or October, depending on the water availability and other factors. On the other hand, there might be crops, such as coconut and banana, that can span an entire year for its cultivation. Thus, instead of using arbitrary sequences as in prior work \cite{pixel-set, pastis-r, sustainbench, cropharvest, agriculture-vision, radiant}, we propose a novel method for preparing time-series data where we consider the length of the regional standard growing season as the duration for creating the time-series data. This strategy leverages the domain knowledge of the crop's growing season in the region, which is available (as a \emph{default}) even in the absence of information on the crop's \emph{actual} growing season in a particular plot, which is often the case in a real-world setting.
For this purpose, we use Table 2 (supplementary material) as a regional standard obtained from Tamil Nadu Agricultural University (TNAU) \cite{tnau}. We then curate a data set based on this regional standard for paddy crop cultivation. Specifically, for each regional standard season, we consider all the satellite images of a particular plot between that duration as a single sample. From an application standpoint, we argue that leveraging domain knowledge is crucial for success.

\textbf{Dataset Creation:} As mentioned in section \ref{sec:ann_technique}, instead of clipping satellite images at a plot level, we clipped a patch of $320m\times320m$ around each plot, which might also contain other plots from the region. Following the technique we proposed in the previous section, we downloaded and clipped the images available in the regional standard growing season only.
\footnote{The dataset directory structure is explained in supplementary material.}


\begin{figure}
    \centering
    \begin{subfigure}[b]{0.32\linewidth}
         \centering
         \includegraphics[width=\textwidth, height=\textwidth]{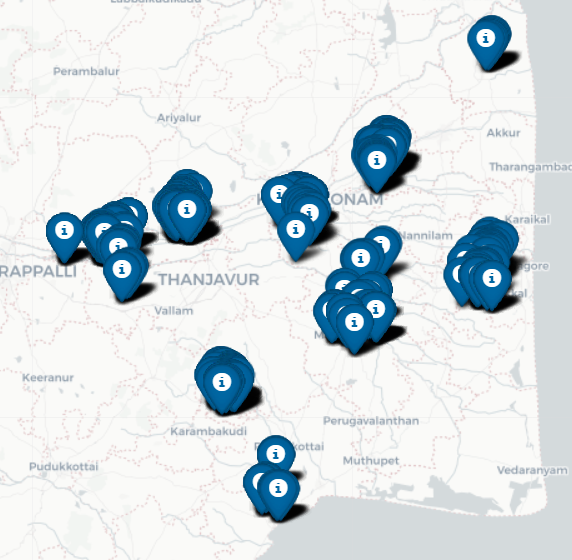}
         \caption{Training plots}
         \label{fig:train-plots}
     \end{subfigure}
    \begin{subfigure}[b]{0.32\linewidth}
         \centering
         \includegraphics[width=\textwidth, height=\textwidth]{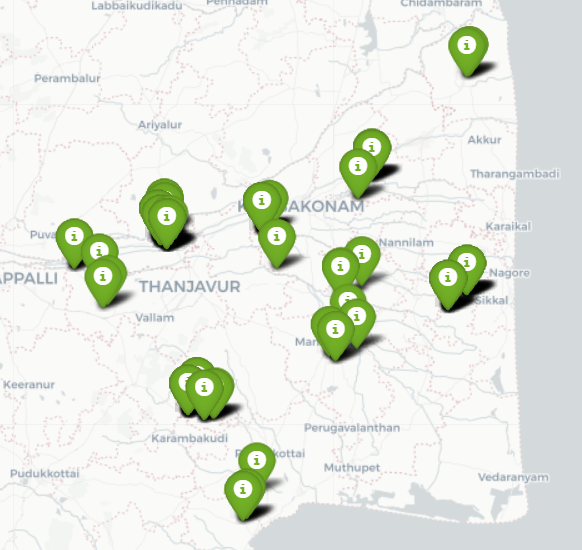}
         \caption{Validation plots}
         \label{fig:val-plots}
     \end{subfigure}
    \begin{subfigure}[b]{0.32\linewidth}
         \centering
         \includegraphics[width=\textwidth, height=\textwidth]{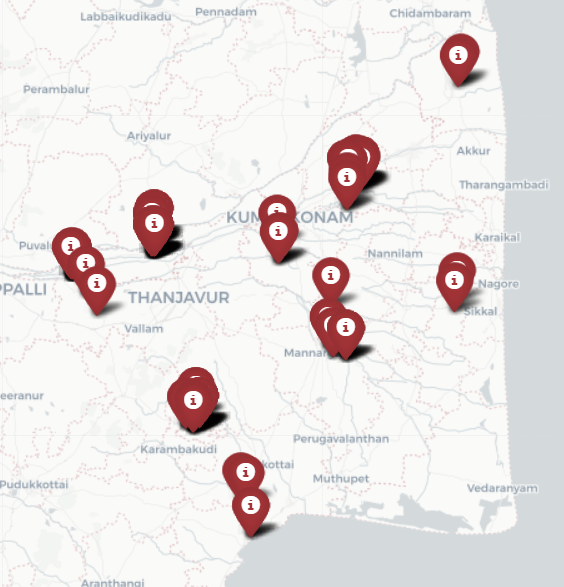}
         \caption{Testing plots}
         \label{fig:test-plots}
     \end{subfigure}
    \caption{Geographical extent of the surveyed region. Each marker denotes the presence of a surveyed plot. The blue color represents plots in the train set, the green color represents the validation set and the red color represents the test set.}
    \label{fig:study_region}
\end{figure}

\textbf{Creating Train/Val/Test Splits:}
\label{sec:split}
Randomly dividing the plots into training, validation and test sets created uneven distributions across these splits. We observed that the distribution is highly skewed in terms of almost all the key parameters when using random splitting. Some of the possible reasons for this could be the sensitivity of crop parameters on geographical location, uneven distribution of size and count of the plots across different patches, and unavailability of all the parameters for some samples. Thus, we created plot-level parameter histograms using pixel-wise annotations of all the samples associated with each plot; and then concatenated all the parameter histograms to create a 1-dimensional histogram for each plot in the dataset. We then iteratively split the plots using a stratified split with different random seeds. For each iteration, we compute the Wasserstein metric (Earth Mover's Distance) between the training, validation and test plot histograms. We executed this method for over $200,000$ iterations and chose the split with the least Mean Wasserstein Distance (Fig \ref{fig:study_region}).

\section{Challenges}
\label{sec:challenge}
Along with the applicability of \dataset{} for developing better algorithms to predict multiple key cropping parameters, we also aim to bridge the gap between the agriculture, RS and ML communities. The proposed dataset has different modalities of raw data from distinct satellites that would allow ML engineers to leverage data-driven methods as well as domain-specific knowledge, which could be exploited to develop practically usable systems. 

\begin{enumerate}
    \item Currently, the publicly available datasets ignore optical images having clouds based on a certain threshold. But paddy is also grown during rainy seasons, thus will contain heavy cloud covers for such samples. When the images corresponding to these samples are ignored during the development phase of a solution, the inferences generated after the deployment will be unanticipated. It is also important to know that including such images will not contribute much towards the solution's performance because of the limitations of optical sensors. However, \dataset{} is also encapsulated with radar observations, which are not sensitive to clouds and other atmospheric interference. Thus sophisticated methods for fusing multiple sensory observations is a necessity.
    \item The annotations masks are based on the ground surveys conducted during January $2018$ - March $2018$. Compared to the crop-cutting based data collection, conducting ground-based surveys is economically more feasible, but the ground truths are much noisier. Thus, considering these annotations as ground truths would be logically incorrect. Alternatively, one can assume these annotations to be weak signals about the ground truth information and thus develop algorithms based on weak supervision.
    \item Majority of the plots in this study region are small farms, with more than $95\%$ plots having an area of less than or equal to $1$ acre and an average size of $0.38$ acres. Thus low-resolution crop parameter masks tend to smoothen the edges of individual plots (Fig \ref{fig:plots}). Moreover, the masks at $30$m resolution are unavailable for very small plots (Fig \ref{fig:small-plots}). Sometimes, small plots adjacent to each other are merged due to the resolution (Fig \ref{fig:multi-plots}). The availability of high-resolution masks allows us to develop methods for generating high-resolution inferences. Figures \ref{fig:plots}, \ref{fig:small-plots} and \ref{fig:multi-plots} depict the boundary and area of individual surveyed plots.

    \begin{figure}
        \centering
        \begin{subfigure}[b]{0.3\linewidth}
            \centering
            \includegraphics[width=\textwidth, height=\textwidth]{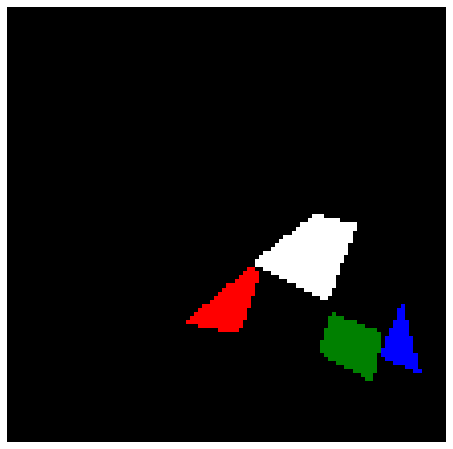}
            \caption{$3$m resolution}
            \label{fig:plot-3}
        \end{subfigure}
        \begin{subfigure}[b]{0.3\linewidth}
            \centering
            \includegraphics[width=\textwidth, height=\textwidth]{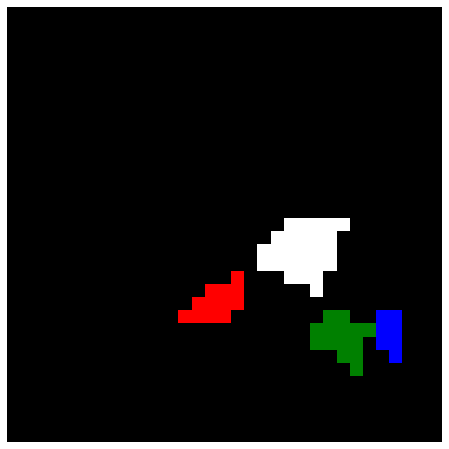}
            \caption{$10$m resolution}
            \label{fig:plot-10}
        \end{subfigure}
        \begin{subfigure}[b]{0.3\linewidth}
            \centering
            \includegraphics[width=\textwidth, height=\textwidth]{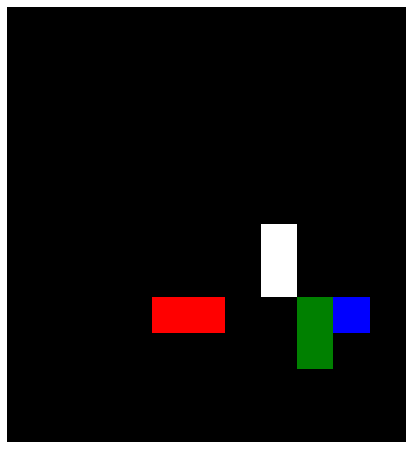}
            \caption{$30$m resolution}
            \label{fig:plot-30}
        \end{subfigure}
        \caption{Masks at LR tend to smoothen the plot boundaries.}
        \label{fig:plots}
    \end{figure}
    
    \begin{figure}
        \centering
        \begin{subfigure}[b]{0.3\linewidth}
            \centering
            \includegraphics[width=\textwidth, height=\textwidth]{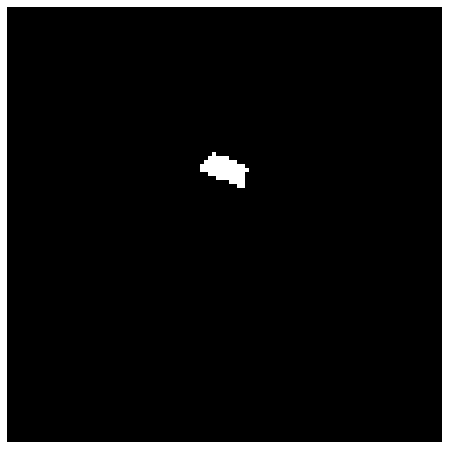}
            \caption{$3$m resolution}
            \label{fig:small-plot-3}
        \end{subfigure}
        \begin{subfigure}[b]{0.3\linewidth}
            \centering
            \includegraphics[width=\textwidth, height=\textwidth]{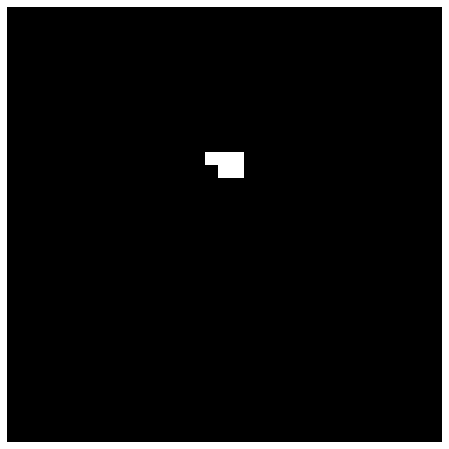}
            \caption{$10$m resolution}
            \label{fig:small-plot-10}
        \end{subfigure}
        \begin{subfigure}[b]{0.3\linewidth}
            \centering
            \includegraphics[width=\textwidth, height=\textwidth]{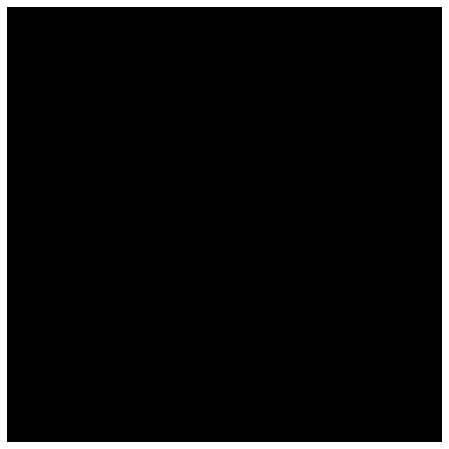}
            \caption{$30$m resolution}
            \label{fig:small-plot-30}
        \end{subfigure}
        \caption{Masks are not available for smallholding farms at LR.}
        \label{fig:small-plots}
    \end{figure}
    
    \begin{figure}
        \centering
        \begin{subfigure}[b]{0.3\linewidth}
            \centering
            \includegraphics[width=\textwidth, height=\textwidth]{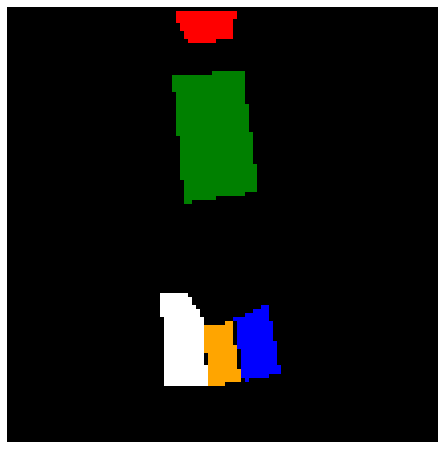}
            \caption{$3$m resolution}
            \label{fig:multi-plot-3}
        \end{subfigure}
        \begin{subfigure}[b]{0.3\linewidth}
            \centering
            \includegraphics[width=\textwidth, height=\textwidth]{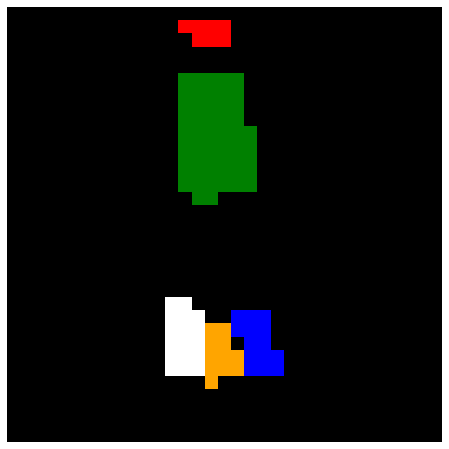}
            \caption{$10$m resolution}
            \label{fig:multi-plot-10}
        \end{subfigure}
        \begin{subfigure}[b]{0.3\linewidth}
            \centering
            \includegraphics[width=\textwidth, height=\textwidth]{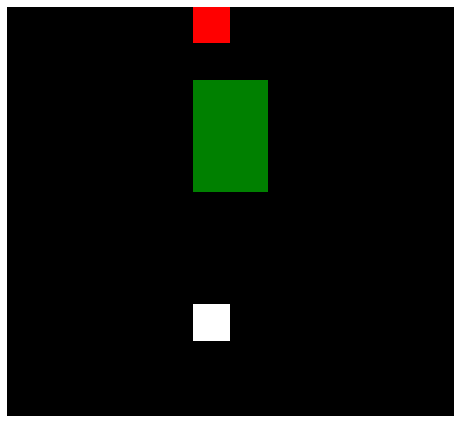}
            \caption{$30$m resolution}
            \label{fig:multi-plot-30}
        \end{subfigure}
        \caption{Masks for small adjacent plots are merged at LR.}
        \label{fig:multi-plots}
    \end{figure}
    
    \item Despite conducting an extensive ground survey, most of the area in the study region remains unsurveyed. Thus the number of data samples available over a season for predicting crop parameters is small. This restriction inspires small-sample and data-efficient learning methods.
    \item Although the sampling strategy allowed us to gather data from all over the study region, there are several other factors that influence the distribution of crop parameters. Factors such as distance from the irrigation source, soil conditions, etc. can vary across different plots and may also influence the prediction accuracy. The predictive models would be required to factor in these changes or be designed to be robust to them in order to be deployed. 
\end{enumerate}

\section{Methodology} \label{sec:methods}
Although \dataset{} can be used to perform various tasks, such as synthetic band generation, image super-resolution, multi-task learning, etc., we benchmark it on $3$ tasks, i.e. crop segmentation, phenology date and crop yield prediction. Phenology date prediction is further composed of sowing, transplanting and harvesting date predictions.
We benchmark the dataset using the code repository of \cite{garnot2021panoptic}, which includes the implementation of U-Net 3D, ConvLSTM and U-TAE among others. We slightly modified the backbone architecture of all the models to output a pixel-wise embedding, instead of a class map, and add a convolution output layer. This modification allows us to perform cross-satellite fusion. 

\subsection{Crop Segmentation}
We pose this as a binary semantic segmentation problem of classifying each pixel as paddy or non-paddy. But one can also pose this as a multi-class semantic segmentation problem as finer labels for all the non-paddy crops are also provided.
We use Pytorch's implementation of the Cross-Entropy Loss. Therefore we don't need to use any activation function explicitly on the logits. This makes the architecture reusable for regression tasks, discussed in the next sections.
Along with the time-series crop segmentation, we also demonstrate the performance of U-Net 2D and DeepLabV3+ using a single image as the input.

\subsection{Phenology Dates Prediction} \label{sec:dateprediction}
Phenology date prediction can be posed as a segmentation as well as a regression problem. We know the duration of regional standard season a priori; hence, we can classify one of the season days as the sowing, transplanting or harvesting day. A common approach in such a problem setting is to use a cross-entropy loss, which is not ideal in this scenario because it treats all the misclassifications equally; hence a date that is misclassified by $\pm1$ day will be the same as a misclassification of $\pm100$ days. Therefore, we pose this problem as a regression problem and use the root mean squared error (RMSE) loss. A regressor also makes it possible to account for outliers, as mentioned in section \ref{sec:challenge}. Because the architecture discussed in the previous section does not contain any activation function at the output layer, it is directly reusable for regression tasks. 

\subsection{Crop Yield Prediction}
\label{sec:crop-yield}
Similar to \ref{sec:dateprediction}, we pose yield prediction as a regression task and use the RMSE loss. We emphasize that even after incorporating the regional standards, it might be challenging to estimate the yield if multiple vegetation signals are available for a pixel. Let's consider a case where the start of the regional standard season contains the harvesting period of a crop $A$ grown in the previous season and the actual growing season for the crop $B$ starts just after the harvesting period of $A$. Because there are multiple vegetation signals within the standard season, the estimates will be ambiguous. If $B$ turns out to be a damaged crop but $A$ has high productivity, the yield estimates will be unanticipated and incorrect. Therefore, we propose a novel strategy of using the phenology dates to estimate the \textit{actual growing season}, which is then used for yield prediction. 
This follows two implementations. One can argue that time-series data is necessary to monitor productivity throughout the season. One can also claim that a single image of the crop can estimate crop yield during the harvest season. Therefore, we benchmark the results of crop yield prediction on both these implementations. 



\subsection{Cross-Satellite Fusion}
\label{sec:fusion}
\begin{figure}[h]
    \centering
    \includegraphics[width=\linewidth]{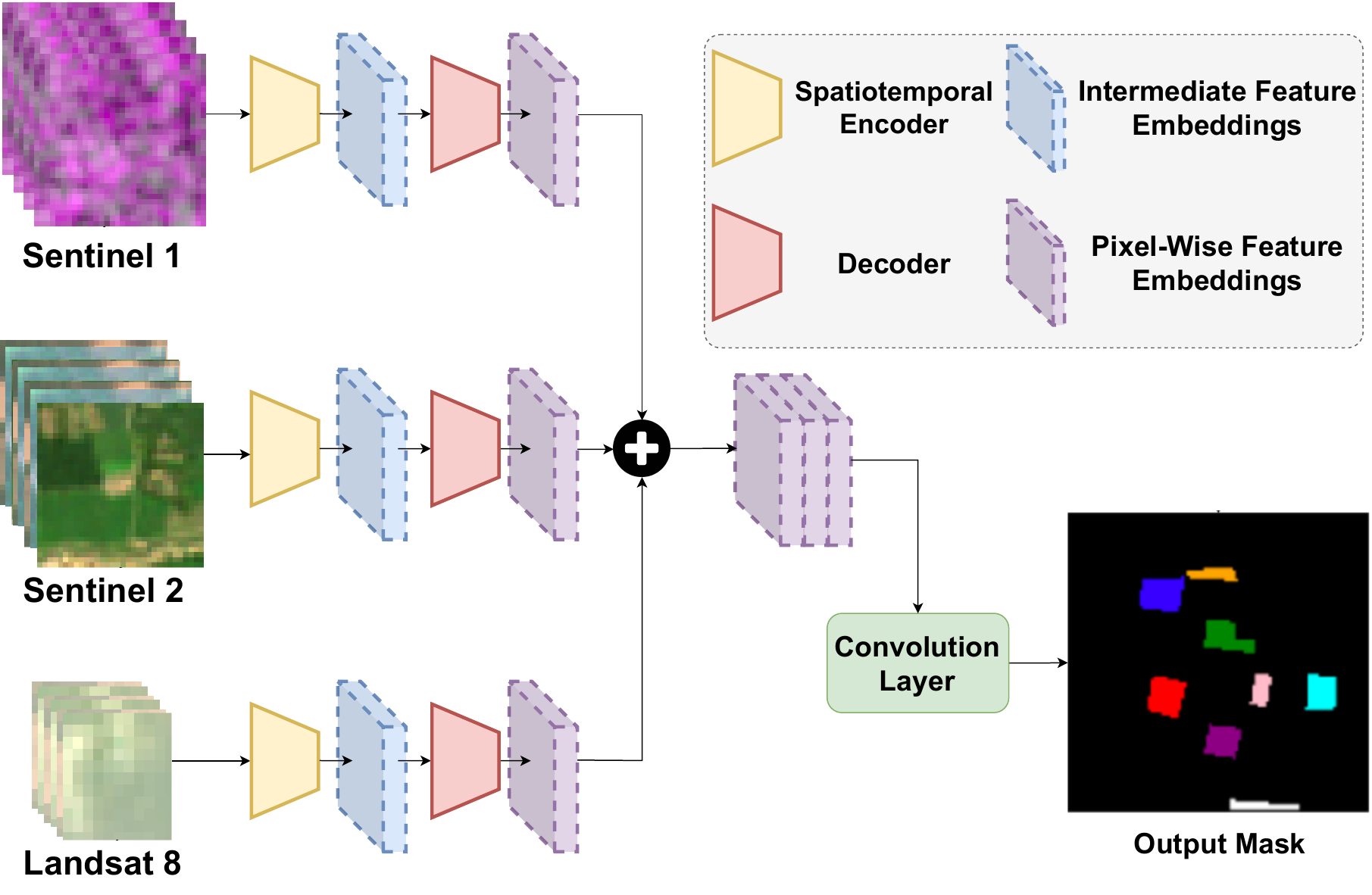}
    \caption{We adopted this architecture to benchmark the dataset on all the tasks. The time-series sequence from each participating satellite is processed using a spatiotemporal encoder to generate an intermediate feature embedding. The decoder accepts these embeddings and generates a pixel-wise embedding. All the pixel-wise embeddings from all the participating satellites are channel-wise concatenated and fed to a convolution output layer. If only one satellite is participating in fusion, it is equivalent to a single satellite prediction task.}
    \label{fig:fusion}
\end{figure}

The data from different satellites can be fused in various ways. The prominent fusion techniques are early fusion: the inputs are concatenated and then processed, and late fusion: the embeddings from each satellite data are concatenated and further processed.
Because the satellites have different revisiting frequencies and hence acquire images on different dates, making it inappropriate to do an early fusion. Therefore, we adopt the late fusion technique as shown in Fig~\ref{fig:fusion}, wherein we use the architectures mentioned in the previous sections to generate a feature embedding instead of a class map. We concatenate the embeddings of all the participating satellites channel-wise and then use a convolution output layer.

\section{Experiments}
\subsection{Experimental Setup}
We use all the \ninstances{} samples for benchmarking crop segmentation. Although rough estimates of the phenology dates are available for all the \npaddy{} paddy samples, we benchmark phenology date and yield estimation on \nyieldinstances{} paddy samples for which all the other crop parameters are also available. We divided the samples into $80\%$, $10\%$ and $10\%$ splits to create training, validation and test sets using the technique mentioned in section \ref{sec:split}. Although, while making the splits we ensured that no plot from the validation and testing sets is exposed during the training, it was observed that plots in validation and test sets were still being exposed during training because some of these plots are adjacent to the plots in the training set. Therefore, by masking them out, we ensure that such pixels are ignored while training. We use a $50\%$ dropout strategy to train all the models for $100$ epochs with a cosine anneal scheduler having starting and minimum learning rate of $0.1$ and $1e-4$, respectively, which is kept constant after $75$ epochs. We set the starting learning rate for all the single image tasks to be $1e-3$. We reduced the number of parameters by reducing the number of convolution blocks because of the small size of the dataset as compared to \cite{garnot2021panoptic}. We also adopted data augmentation techniques, which included random horizontal and vertical flip along with random brightness adjustment and Gaussian blur. While the flipping augmentation helps in better spatial regularization of the plots, brightness and blur augmentations help by synthesizing the effect of climatic and atmospheric interference. Moreover, because there is an imbalance of crop type distribution with paddy being less, we use a weighted cross-entropy loss with weights $0.62013$ and $0.37987$ for paddy and non-paddy crops, respectively. For a fair comparison between the reported results, we resize the Landsat $8$ images to generate predictions at $10m$ resolution.
\begin{table*}[!ht]
    \centering
    \begin{tabular}{l|c|c|c|c|c}
        \toprule
        \textbf{Task} & \textbf{Metric} & \textbf{L8} & \textbf{S2} & \textbf{S1} & \textbf{Fusion} \\
        \midrule
        Crop Type (SI)  &  IoU (\%)     & $47.73\% \pm 1.77\%$                          & $54.87\% \pm 3.08\%$         & \cellcolor{\rcolor}\textbf{$\textbf{64.35\%} \pm \textbf{4.82\%}$}                         & -                   \\
        Crop Type  & IoU (\%)     & $56.04\% \pm 5.84\%$                          & $78.12\% \pm 3.48\%$         & \cellcolor{\rcolor}$\textbf{81.77\%} \pm \textbf{6.60\%}$                         & $\textbf{81.07\%} \pm \textbf{5.77\%}$             \\
        Sow Date                      & MAE (days)      & $2.66 \pm 0.961$                            & \cellcolor{\rcolor}$\textbf{2.30} \pm \textbf{0.611}$           & $3.61 \pm 0.898$                            & $\textbf{2.33} \pm \textbf{0.639}$              \\
        Transplant Date               & MAE (days)      & $\textbf{6.20} \pm \textbf{1.030}$                            & $6.36 \pm 2.164$           & $7.23 \pm 0.779$                           & \cellcolor{\rcolor}$\textbf{6.16} \pm \textbf{1.770}$               \\
        Harvest Date                  & MAE (days)      & $\textbf{9.86} \pm \textbf{0.736}$                           & \cellcolor{\rcolor}$\textbf{8.83} \pm \textbf{1.520}$           & $10.08 \pm 0.561$                           & $10.75 \pm 3.389$               \\
        Crop Yield (SI)           & MAPE (\%)       & \cellcolor{\rcolor}$\textbf{46.74\%} \pm \textbf{3.82\%}\%$                          & $60.44\% \pm 14.50\%$         & $\textbf{48.35\%} \pm \textbf{7.64\%}$                         & -                   \\
        Crop Yield (RS)   & MAPE (\%)       & \cellcolor{\rcolor}$\textbf{54.00\%} \pm \textbf{9.67\%}$  &                       $72.38\% \pm 8.74\%$         & $71.81\% \pm 17.27\%$                           & $\textbf{70.35\%} \pm \textbf{13.75\%}$             \\
        Crop Yield (AS)       & MAPE (\%)       & \cellcolor{\rcolor} $\textbf{59.38\%} \pm \textbf{14.75\%}$ &                       $73.59\% \pm 9.81\%$         & $65.66\% \pm 16.24\%$                         & $\textbf{64.56\%} \pm \textbf{13.77\%}$ \\
        \bottomrule
    \end{tabular}
    \caption{Results for the benchmarking tasks. Single-image experiments are denoted with SI in parenthesis. The results are reported using the same benchmarking model (U-Net 3D for time-series and U-Net 2D for single image) for a fair comparison. RS denotes the experiment when using Regional Standards to create the time-series input, whereas AS denotes the one using Actual Season. The supplementary material includes a list of exhaustive experiments and their results. }
    \label{tab:results}
\end{table*}

\begin{figure*}
    \centering
    \includegraphics[width=\textwidth]{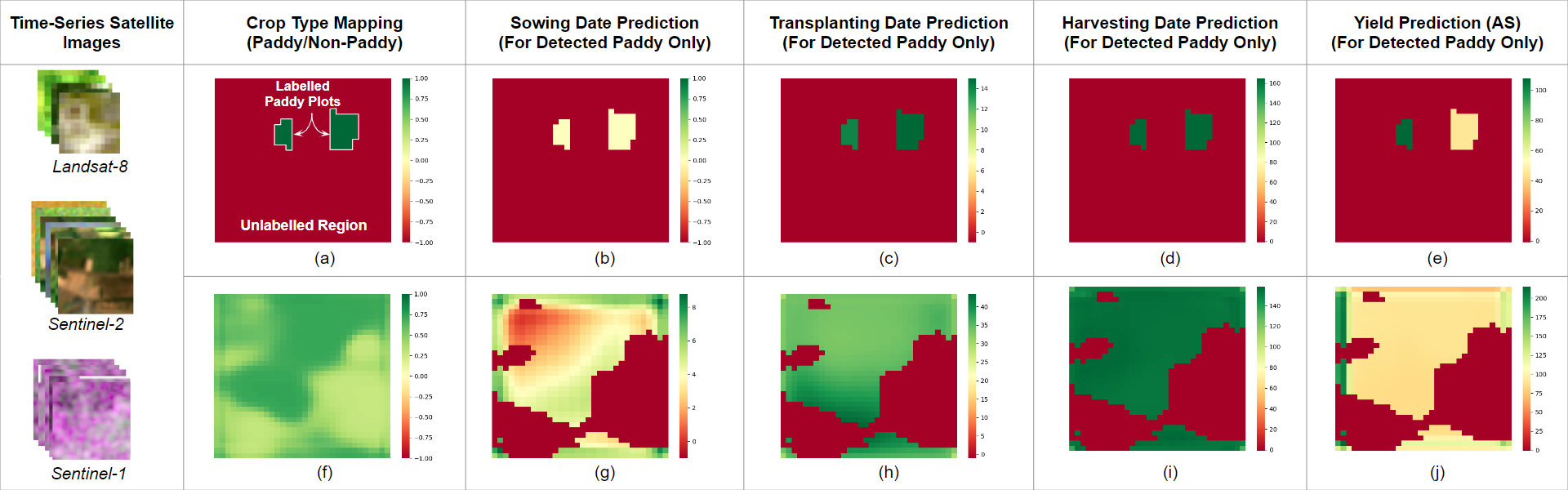}
    \caption{An illustration of the outputs obtained from U-Net 3D (Fusion) on the test set. Figures (a) - (e) are the ground truth annotations and figures (f) - (j) are the output maps generated. All the pixels that were predicted as non-paddy in (f) were masked out in (g) - (j).}
    \label{fig:enter-label}
\end{figure*}

\subsection{Metric}
We report the mean and standard deviation of the results obtained with different seed values (ranging from $0$ to $4$) on the test set. For segmentation tasks, we report the results using pixel-wise accuracy, f1-score and intersection-over-union (IoU). The results include macro-averaged as well class-wise metrics. For regression tasks, we report the root mean squared error (RMSE), mean absolute error (MAE) and mean absolute percentage error (MAPE). Because the phenology date prediction has a specified range of dates, we normalize the MAE by the maximum possible date, i.e. $183$, instead of the ground truth value to compute MAPE, because otherwise, its value overshoots due to the minimal range of sowing date ($\approx0$).

\subsection{Results}
We performed experiments with multiple state-of-the-art architectures for time series and single-image predictions wherever applicable. Table \ref{tab:results} includes the benchmarking results for each task. It is important to note that these results only have the experiments done on the same backbone architecture for a fair comparison. A complete list of experiments is available in the supplementary material. The best results are highlighted in \colorbox{\rcolor}{\textbf{green}} color and the competing (second best) are represented using a \textbf{bold font}. 

\section{Conclusion}
This paper introduces \dataset, a first-of-its-kind dataset consisting of satellite image time-series data from multiple satellites with multiple key cropping parameters annotated at multiple resolutions. The dataset enables researchers to develop an end-to-end pipeline wherein one can predict the key cropping parameters at a plot-level given the satellite images of the complete region. Such an end-to-end pipeline is necessary for real-world deployments where gathering the individual plot-level insights is a crucial task.

Besides the cropping parameter annotations, the dataset also serves as the first step towards building a research ecosystem for researchers across agriculture, remote sensing and machine learning domains by emphasizing on various fundamental challenges that could potentially benefit the overall development of the community.

\section*{Acknowledgement}
Depanshu Sani was supported by Google’s AI for Social Good “Impact Scholars” program, 2021. Saket Anand gratefully acknowledges for the partial support from the Infosys Center for Artificial Intelligence at IIIT-Delhi. We also appreciate Parichya Sirohi's contributions in the early stages of the project. Additionally, we wish to express our gratitude to Dr. Gopinath R. and Dr. Rajakumar R. from Ecotechnology, MS Swaminathan Research Foundation, Chennai, for their valuable inputs concerning the study area and assistance with field data collection.

{\small
\bibliographystyle{ieee_fullname}
\bibliography{egbib}
}

\end{document}


\title{Supplementary Material for ``SICKLE: A Multi-Sensor Satellite Imagery Dataset Annotated with Multiple Key Cropping Parameters"}

\author{Depanshu Sani$^1$, Sandeep Mahato$^2$, Sourabh Saini$^1$, Harsh Kumar Agarwal$^1$, \\
Charu Chandra Devshali$^2$, Saket Anand$^1$, Gaurav Arora$^1$, Thiagarajan Jayaraman$^2$ \\
$^1$Indraprastha Institute of Information Technology, Delhi, India\\
$^2$MS Swaminathan Research Foundation, Chennai, India\\
{\tt\small \url{https://sites.google.com/iiitd.ac.in/sickle/home}}
}
\maketitle

\section{Rationale for Location and Crop Selection}
We study paddy cultivation in the Cauvery Delta, which is considered to be a major paddy cultivation region in Tamil Nadu, India, which supports food security and livelihoods among millions of farmers in the region \cite{climate_change_affect}. In any given agricultural year, the farmers in our study region are primarily dependent on irrigation from surface-water and groundwater sources. Additionally, they rely on two monsoon seasons for rainfall: the Southwest Monsoon, which occurs from June to September, and the Northeast Monsoon, from October to December. Regarding the agricultural practices in the Delta, they are intrinsically aligned with a "water availability" calendar. To illustrate this, in a typical agricultural year, farmers base their decisions on whether to cultivate one or two paddy crops on the availability of water. These decisions are influenced by a variety of factors, including surface and groundwater irrigation, as well as the predictable timing of the Southwest Monsoon (June-September) and Northeast Monsoon (October-December). Therefore, the Delta's agricultural system has been shaped over time to adapt to this water availability calendar.

There's growing evidence that points towards significant shifts in the Delta's agrarian landscape, specifically the stagnation in paddy yields and an uncertainty in water availability for paddy cultivation \cite{cropping_pattern}. An Indian Council for Agricultural Research (ICAR) study in 2013 reclassified the four districts of the Cauvery delta from dry semi-humid to semi-arid conditions \cite{ICAR_report}. In the context of these changes reported for the region and its centrality to rice production, understanding these emerging variations and challenges in paddy cultivation in the Delta becomes crucial.

\section{Spatial Distribution and Sampling Methods}
The Cauvery Delta Zone (CDZ) as shown in Fig \ref{fig:study_region} spans across four principal districts: Thiruvarur, Thanjavur, Nagapattinam, and Mayiladuthurai of Tamil Nadu, India. Collectively, these districts make up $57\%$ of the CDZ. The alluvial soils within this region are prime for wet rice cultivation, complemented by the ancient irrigation system of the Cauvery delta \cite{cropping_pattern}. Integral irrigation structures like the Grand Anicut and the Mettur dam ensure the distribution of water to these districts.

\begin{figure}
    \centering
    \includegraphics[width=\linewidth]{images/Study-Region(Blocks).png}
    \caption{ The study region is part of the Cauvery Delta Zone (CDZ). The sample taluks highlighted in the map are taluks from where field samples were collected for the study.}
    \label{fig:study_region}
\end{figure}

\subsection{Data Collection Blocks}
The blocks for data collection, outlined in Table \ref{tab:blocks}, are sub-district administrative divisions positioned across the downstream parts of the Vennar and Cauvery rivers. Their selection was guided by the criticality of irrigation water availability, influenced by river flows managed by a series of dams and anicuts (Fig \ref{fig:study_region}). The timing of water release and its downstream availability profoundly impacts paddy crop sowing dates across the Delta \cite{palakurichi_study}.

\begin{table*}[h]
  \begin{center}
    {\small{
\begin{tabular}{llll}
\toprule
\textbf{Block} & \textbf{District} & \textbf{Region (by source of irrigation)} & \textbf{Phase} \\
\midrule
Thiruvaiyaru & Thanjavur & Upper Cauvery & Phase One \\
Kordacheri & Thanjavur & Middle Vennar & Phase Two \\
Sethubavasathiram & Thanjavur & Coastal GAC & Phase Two \\
Thiruvidaimaruthur & Thanjavur & Middle Cauvery & Phase Two \\
Thiruvonam & Thanjavur & Coast GAC & Both Phase \\
Valangaiman & Thiruvarur & Middle Vennar & Phase One \\
Mannargudi & Thiruvarur & Middle Vennar & Phase One \\
Kilvelur & Nagapattinam & Coastal Vennar & Both Phase \\
Kuthalam & Mayiladuthurai & Middle Cauvery & Phase One \\
Sirkali & Mayiladuthurai & Coastal Cauvery & Phase Two \\
\bottomrule
\end{tabular}
}}
\end{center}
\caption{List of blocks and the districts where data collection was conducted in the Delta. GAC refers to the Grand Anicut Canal. Upper, Middle and Coastal refer to a rough three-fold division of the length of the Cauvery and Vennar downstream of the Grand Anicut, from where also the GAC originates.}
\label{tab:blocks}
\end{table*}

\subsection{Field Data Collection}
\begin{enumerate}
    \item \textbf{Phase One (2018-2020):} Emphasis on collecting crop-type information at the plot level. This data helps classify paddy and non-paddy plots, detailing various crop types and their sowing and harvesting periods under different seasons.
    \item \textbf{Phase Two (2019-2020):} Emphasis on detailed information on paddy cultivation, specifically yield data, which is instrumental for the calibration and validation of the proposed remote-sensing-based paddy yield prediction model. The collection was limited to two years to reduce the possibility of errors in the yield records.
\end{enumerate}

All data was consolidated from farmer interviews, structured questionnaires, and the Kobo Collect android app.

\section{Directory Structure}
The dataset is organized according to the following directory structure. Considering we want to fetch Landsat-8's image of band B1 acquired on $2018-06-16$ for the $1^{st}$ sample, the path will be \path{L8/1/LC08_142052_20180616/LC08_142052_20180616.SR_B1.tif}, shown in the parenthesis. 

\dirtree{%
.1 \textbf{Satellite Name} (\textit{L8}).
.2 \textbf{Sample ID} (\textit{1}).
.3 \textbf{Date} (\textit{LC08\_142052\_20180616}).
.4 \textbf{BandID.tif} (\textit{LC08\_142052\_20180616.SR\_B1.tif}).
.4 \vdots.
.3 \vdots.
.2 \vdots.
.2 \textbf{Metadata} (\textit{metadata.pkl}) .
}

\section{Dataset Analysis}
\begin{figure*}[b]
    \centering
    \begin{subfigure}[b]{\linewidth}
        \centering
        \includegraphics[width=\linewidth]{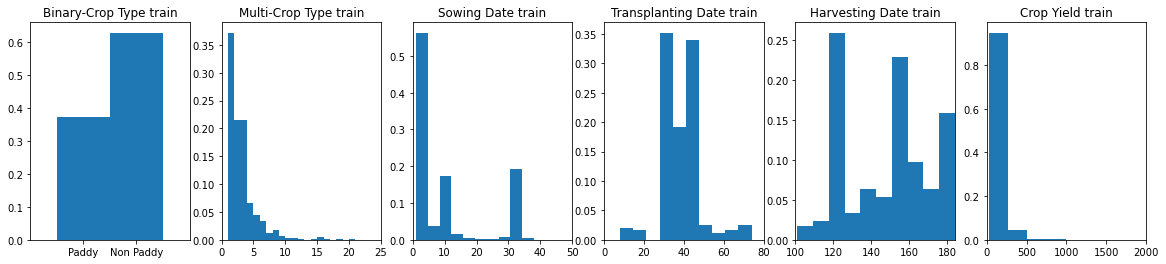}
        \caption{Training Set}
        \label{fig:train-dist}
    \end{subfigure}
    \begin{subfigure}[b]{\linewidth}
        \centering
        \includegraphics[width=\linewidth]{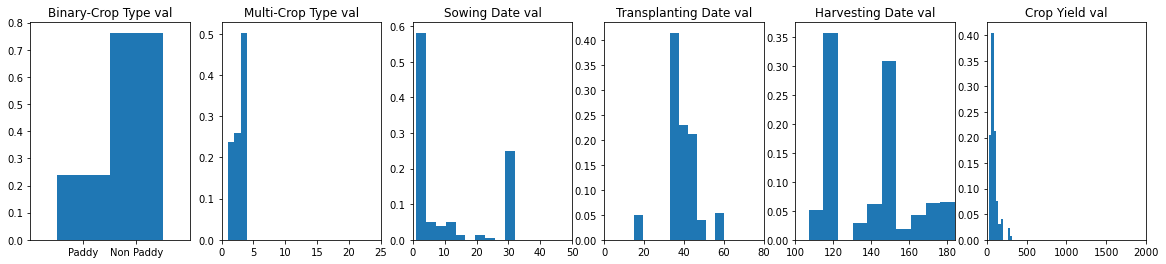}
        \caption{Validation Set}
        \label{fig:val-dist}
    \end{subfigure}
    \begin{subfigure}[b]{\linewidth}
        \centering
        \includegraphics[width=\linewidth]{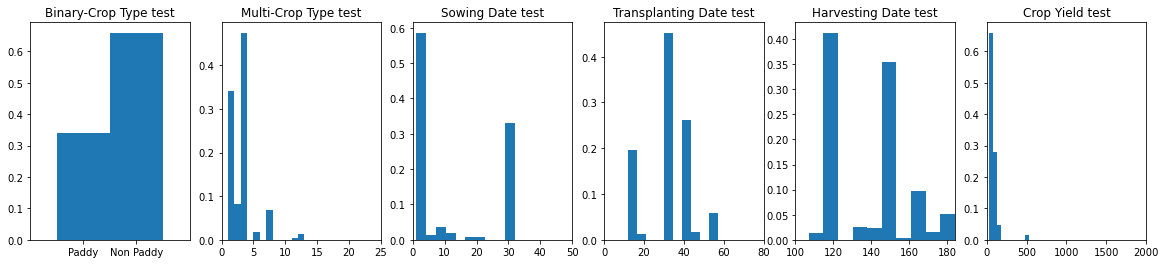}
        \caption{Testing Set}
        \label{fig:test-dist}
    \end{subfigure}
    \caption{\textbf{Data distribution of training, validation and test sets.} Distribution of the pixel-wise annotations for all the key crop parameters using the split based on Wasserstein distance. Here, the Y-axis denotes the percentage of total number of pixels and the X-axis denotes the crop type, phenology dates and crop yield. It can be observed that the distribution of various crop parameters is approximately the same across all the sets using this technique.}
    \label{fig:distribution}
\end{figure*}

\begin{figure*}
    \centering
    \includegraphics[width=\linewidth]{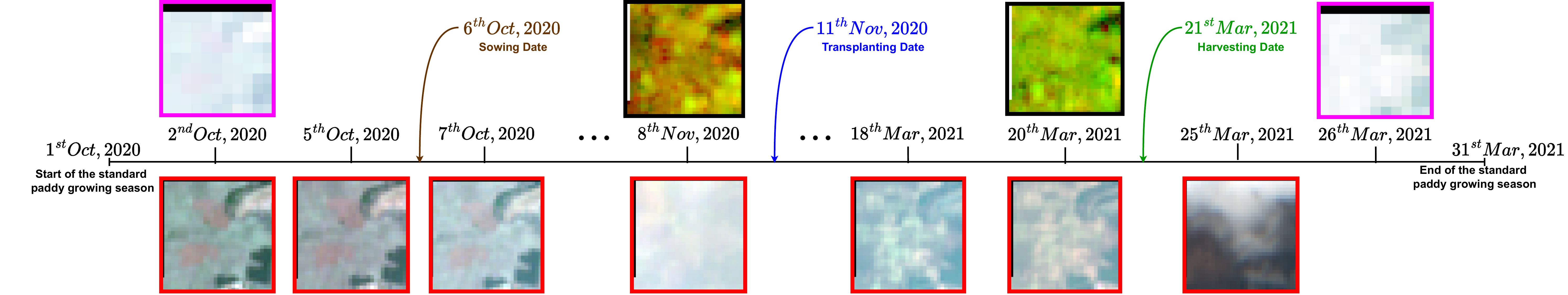}
    \caption{\textbf{Sample input from the dataset.} The axis denotes the individual days from the start of the paddy growing season till the end of that season. The extreme ends of this axis represent the start and the end of the regional standard season. We visualize a few image patches acquired on the corresponding dates from different satellites. Images with pink boundaries denote images acquired from Landsat-8, black boundaries denote Sentinel-1 and red boundaries denote Sentinel-2. Multiple observations from different satellites are observed on various dates. It is also possible to get multiple observations from different satellites on the same day, as shown in the figure. Moreover, as in this case, there may be no images on the sowing, transplanting and/or harvesting dates.}
    \label{fig:sample-images}
\end{figure*}

\begin{figure*}[b]
    \centering
    \begin{subfigure}[b]{0.8\linewidth}
        \centering
        \includegraphics[width=\linewidth]{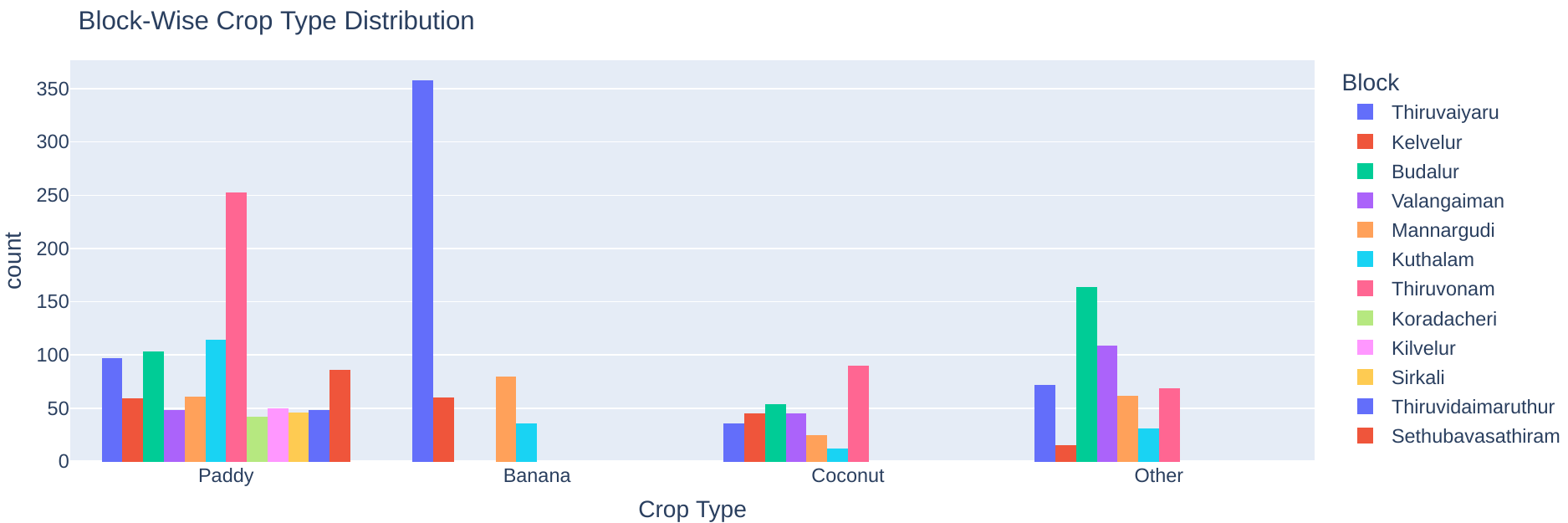}
        \caption{}
        \label{fig:block-crop}
    \end{subfigure}
    \begin{subfigure}[b]{0.8\linewidth}
        \centering
        \includegraphics[width=\linewidth]{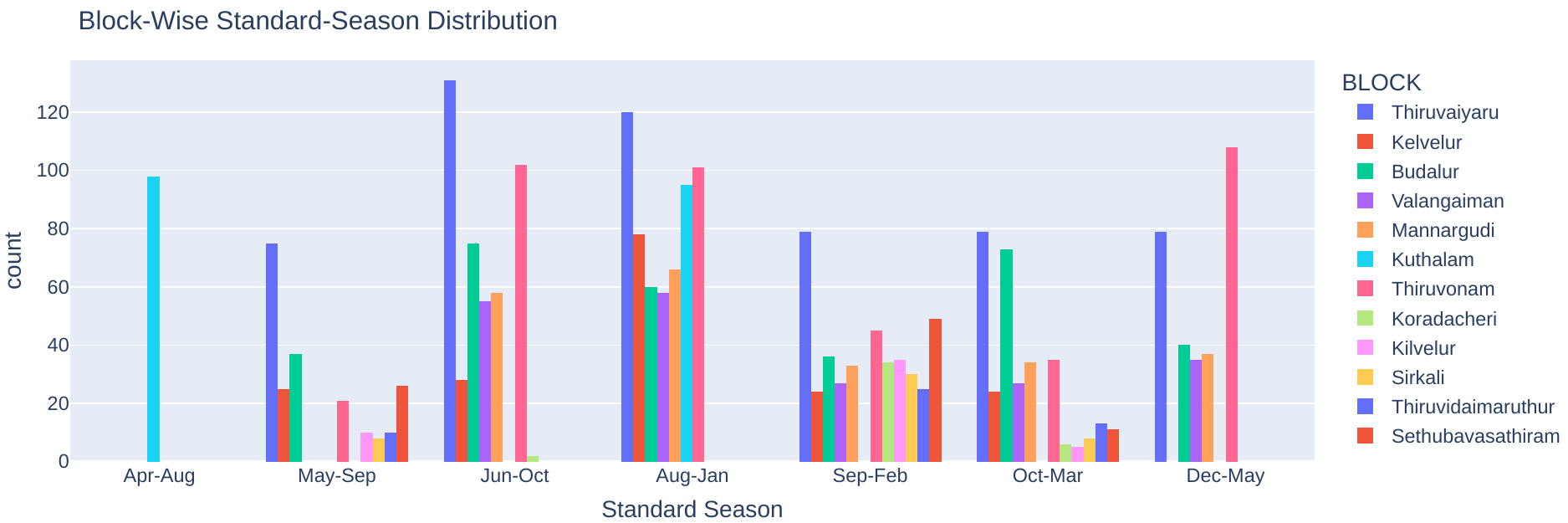}
        \caption{}
        \label{fig:block-seasons}
    \end{subfigure}
    \begin{subfigure}[b]{0.8\linewidth}
        \centering
        \includegraphics[width=\linewidth]{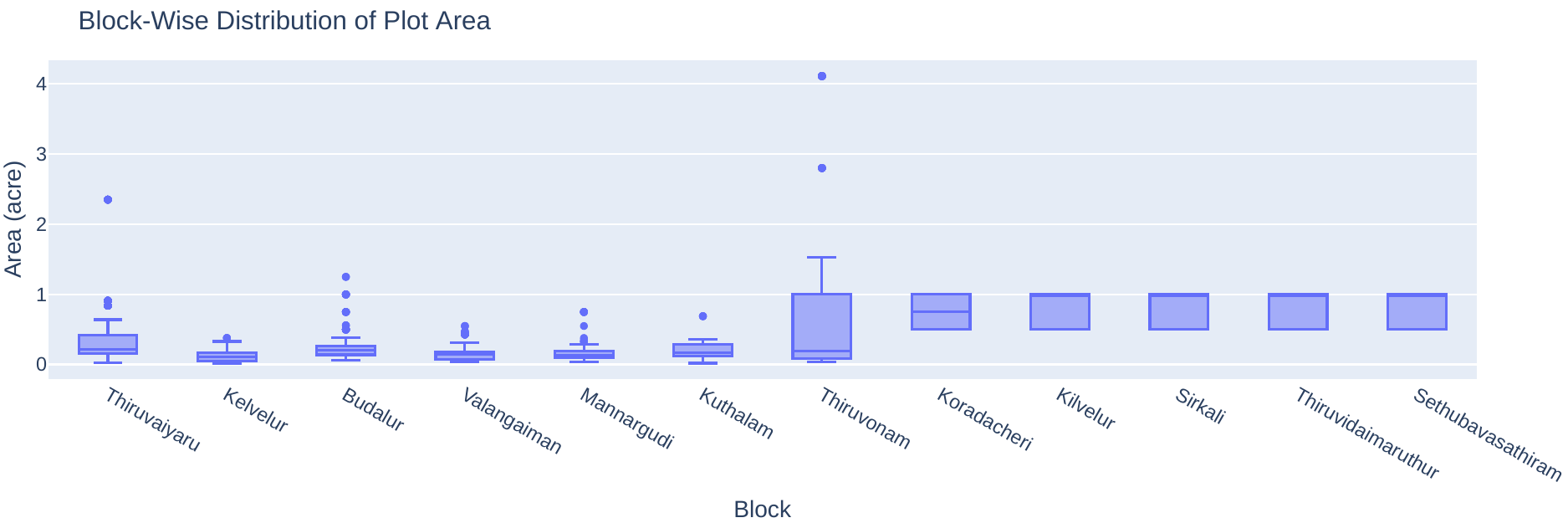}
        \caption{}
        \label{fig:block-area}
    \end{subfigure}
    \begin{subfigure}[b]{0.8\linewidth}
        \centering
        \includegraphics[width=\linewidth]{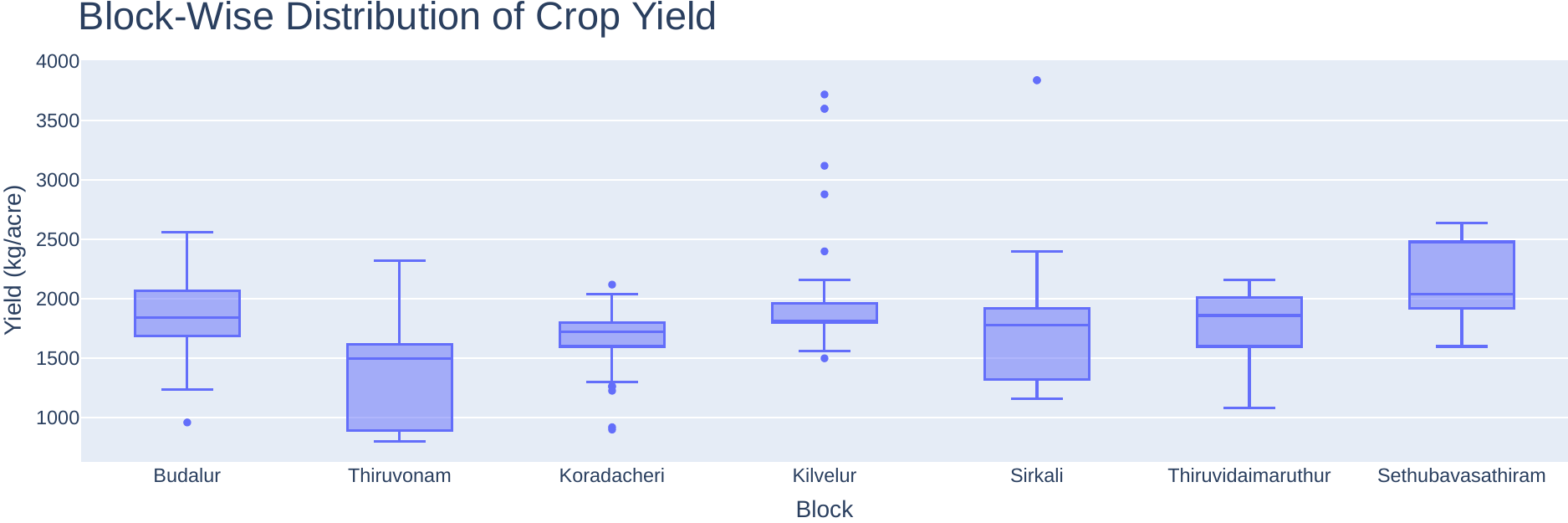}
        \caption{}
        \label{fig:block-yield}
    \end{subfigure}
    \caption{Block-Level Dataset Statistics. Figure \ref{fig:block-crop} shows the distribution of various types of crops in the dataset that are cultivated in the study region in each of the studied blocks. Figure \ref{fig:block-area} represents the area (in acre) of each of the \nplots{} individual plots in the block. Figure \ref{fig:block-seasons} shows the distribution of the standard paddy seasons observed in the study region for each block. Figure \ref{fig:block-yield} shows the block-wise distribution of the crop yield.}
    \label{fig:block-dataset-analysis}
\end{figure*}

\begin{figure*}[b]
    \centering
    \begin{subfigure}[b]{0.48\linewidth}
        \centering
        \includegraphics[width=\linewidth]{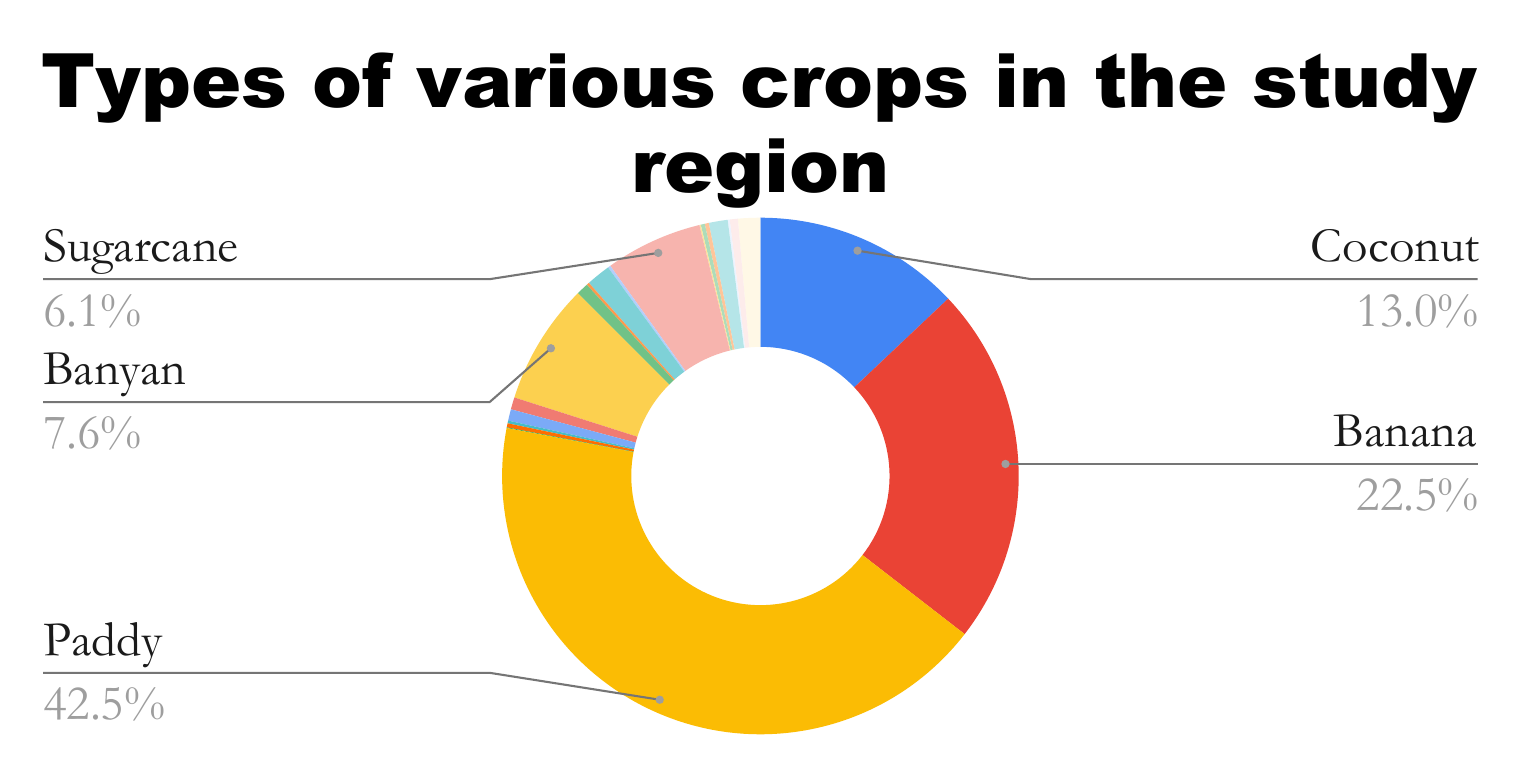}
        \caption{}
        \label{fig:n-crop}
    \end{subfigure}
    \begin{subfigure}[b]{0.48\linewidth}
        \centering
        \includegraphics[width=\linewidth]{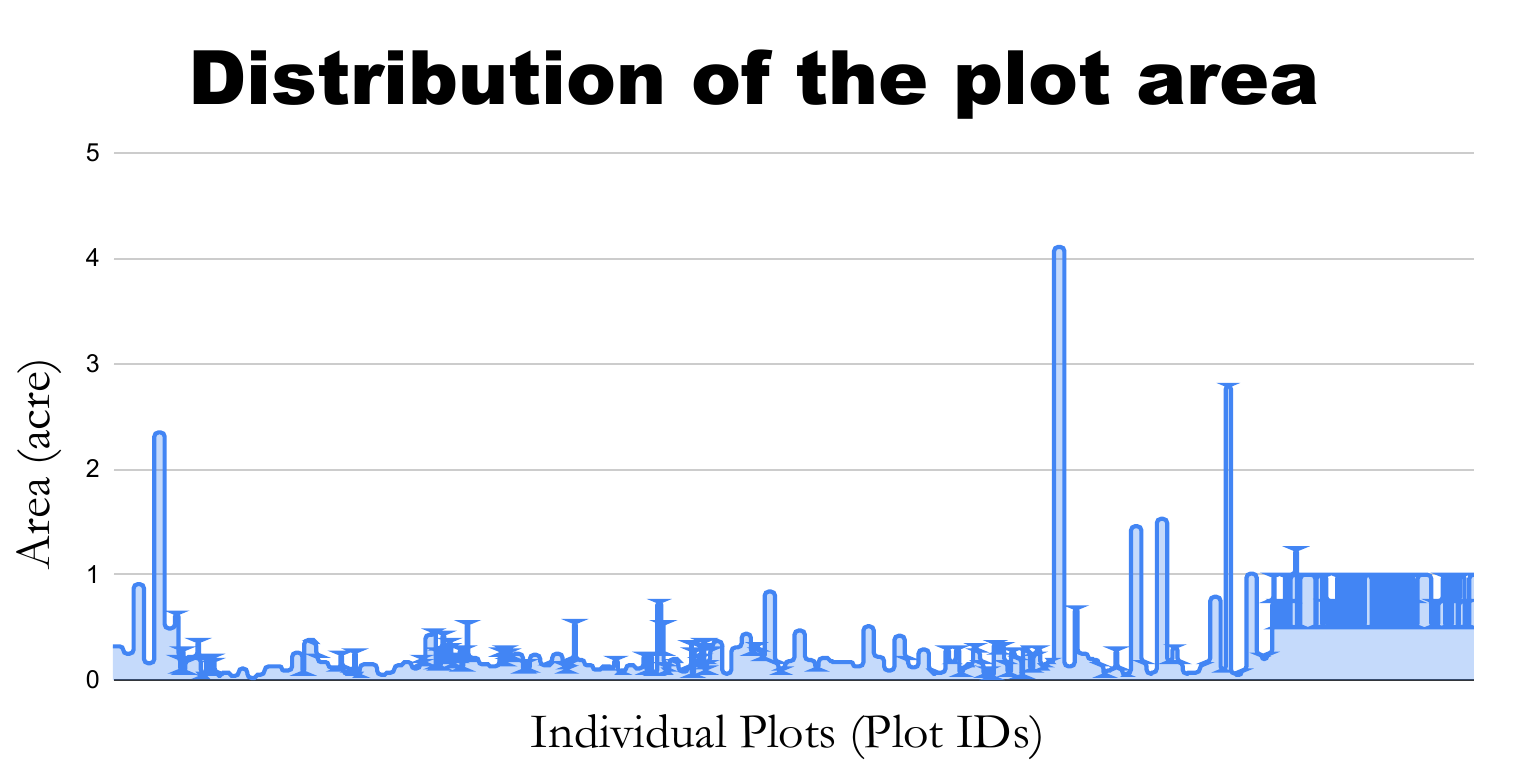}
        \caption{}
        \label{fig:area}
    \end{subfigure}
    \begin{subfigure}[b]{0.48\linewidth}
        \centering
        \includegraphics[width=\linewidth]{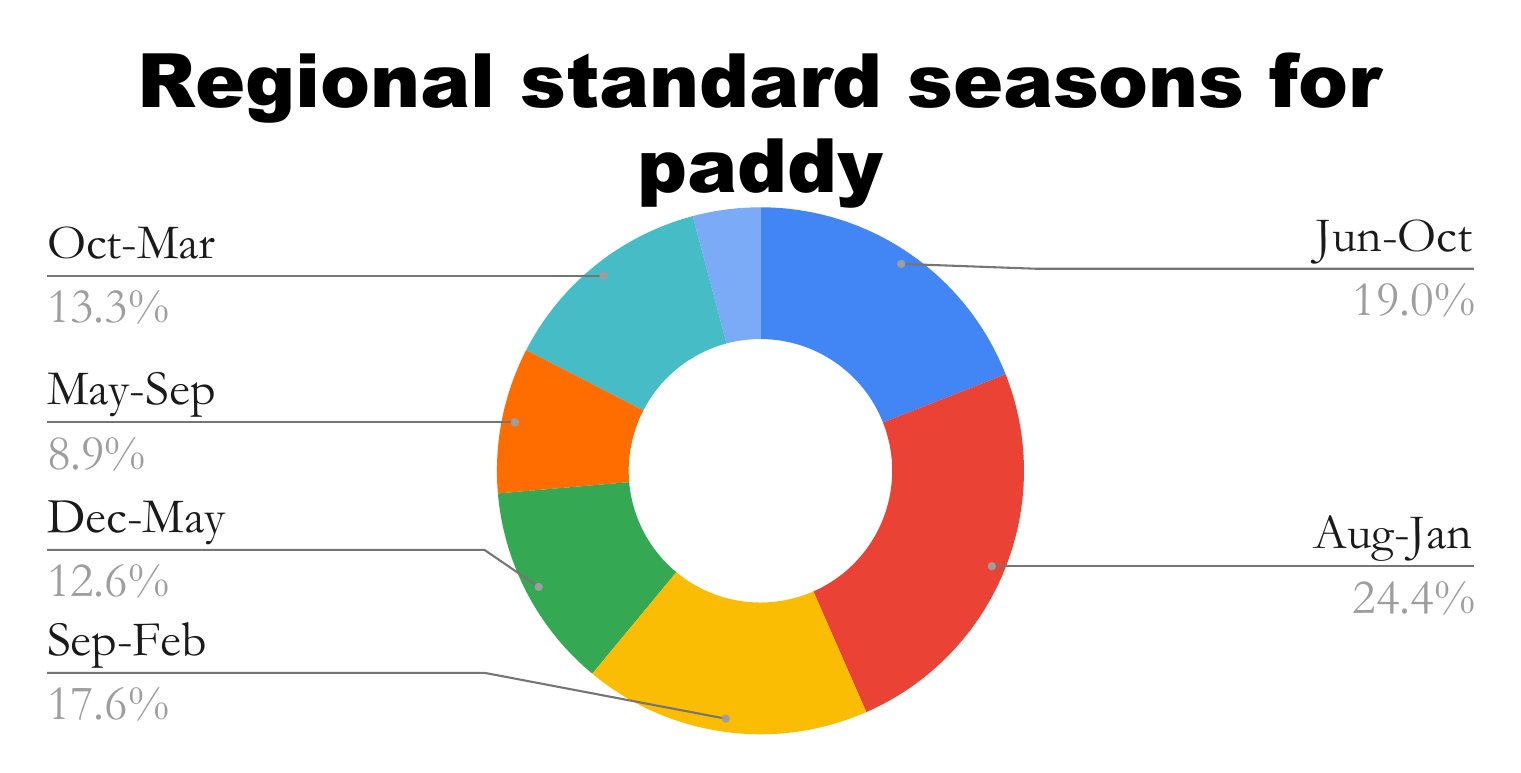}
        \caption{}
        \label{fig:n-seasons}
    \end{subfigure}
    \begin{subfigure}[b]{0.48\linewidth}
        \centering
        \includegraphics[width=\linewidth]{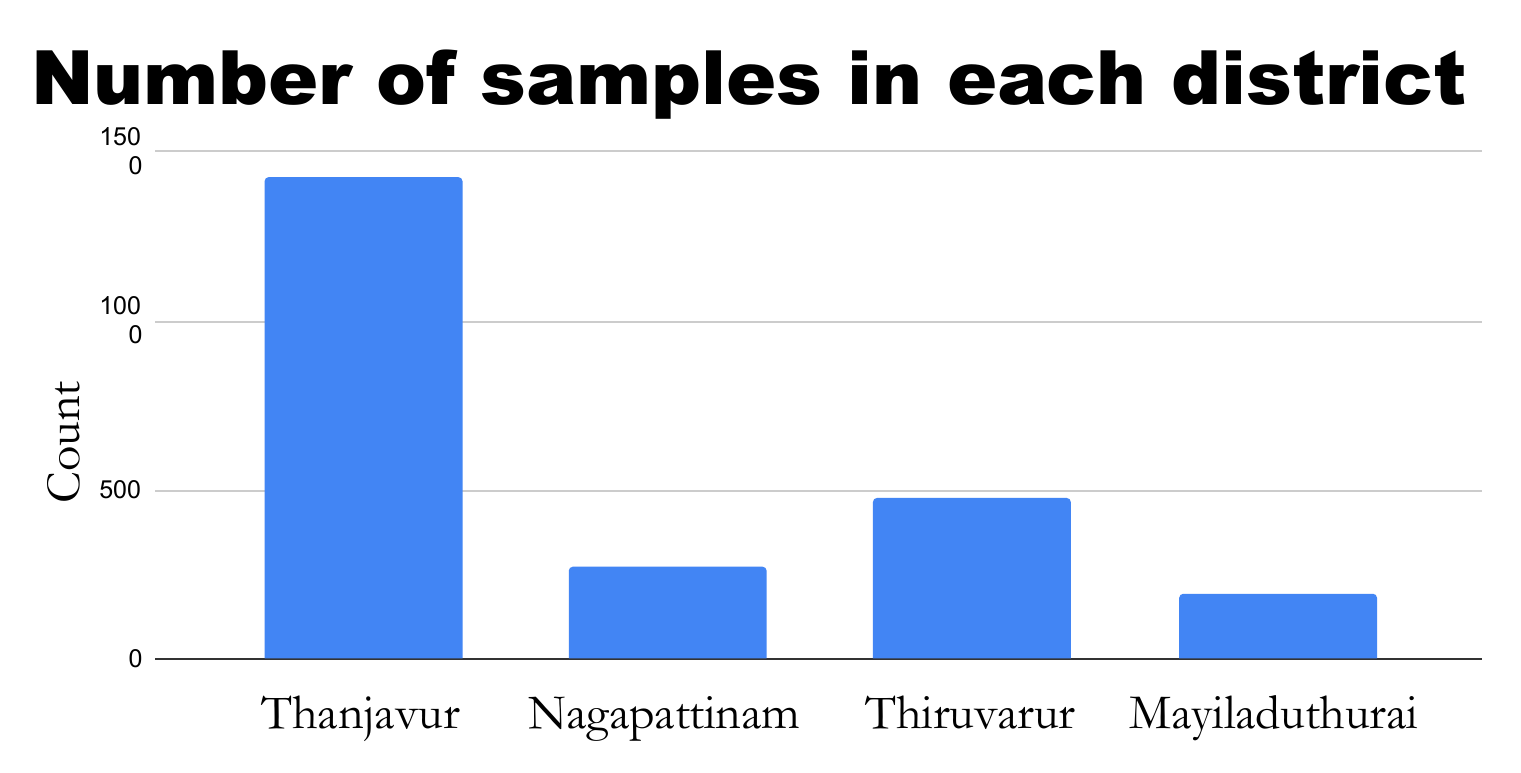}
        \caption{}
        \label{fig:n-samples}
    \end{subfigure}
    \caption{Dataset Statistics. Figure \ref{fig:n-crop} shows the distribution of various types of crops in the dataset that are cultivated in the study region. Figure \ref{fig:area} represents the area (in acre) of each of the \nplots{} individual plots. X-axis denotes the plot id of each plot and Y-axis denotes the area (in acre). Figure \ref{fig:n-seasons} shows distribution of the standard paddy seasons observed in the study region. Figure \ref{fig:n-samples} shows the district-wise distribution of the collected \ninstances{} samples.}
    \label{fig:dataset-analysis}
\end{figure*}

Fig \ref{fig:distribution} presents the pixel-wise key crop parameters' distribution for the dataset using the split strategy based on the Wasserstein distance. Fig \ref{fig:dataset-analysis} presents the statistics of the entire dataset. It includes the various types of crops that are cultivated in the Cauvery Delta region in Tamil Nadu, the area of each plot in acre, regional standard seasons (Table \ref{tab:standardseasons}) present in the dataset that are mapped using \cite{tnau}, and the total number of samples in each district. Fig \ref{fig:block-dataset-analysis} shows a more fine-grained dataset analysis at a block level. Along with other statistics, Fig \ref{fig:block-dataset-analysis} also contains the distribution of crop yield for each block. Figure \ref{fig:sample-images} demonstrates a sample input from the dataset. A few images acquired on the dates mentioned on the axis are visualized in the figure, along with the collected phenology dates. As in this example sample, there are cases where multiple images from different satellites are available on the same day, but no images on the sowing, transplanting and harvesting are available.

\begin{table}[h]
  \begin{center}
    {\small{
\begin{tabular}{lcc}
\toprule
\textbf{Season} & \textbf{Sowing Month} & \textbf{Duration (days)} \\
\midrule
Navarai & Dec. - Jan. & 120 \\
Sornavari & Apr. - May. & 120 \\
Early Kar & Apr. - May. & 120 \\
Kar & May. - June & 120 \\
Kuruvai & June - July & 120 \\
Early Samba & July - Aug. & 135 \\
Samba & Aug. & 180 \\
Late Samba & Sep. - Oct. & 135 \\
Thaladi & Sep. - Oct. & 135 \\
Late Pishanam & Sep. - Oct. & 135 \\
Late Thaladi & Oct. - Nov. & 120 \\
\bottomrule
\end{tabular}
}}
\end{center}
\caption{Regional standard growing seasons for paddy cultivation in the Cauvery Delta region obtained from Tamil Nadu Agricultural University (TNAU) \cite{tnau}. The time-series dataset is created using the maximum duration of the growing paddy season.}
\label{tab:standardseasons}
\end{table}

\section{Additional Experiments and Results}
Tables \ref{tab:unet2d-ct} - \ref{tab:utae-cy-as} include benchmarking results for various state-of-the-art methods on single and time-series image prediction tasks. For single-image tasks, we present the results using U-Net 2D and DeepLabV3+, and for time-series tasks, we present the results using U-Net 3D, ConvLSTM and U-TAE. 
Fusion was not performed for single-image tasks because it is unlikely to capture images from multiple satellites on the same day (or within a small time window) because of their low revisit frequency. Crop signatures change significantly within a few days, and hence, it is not ideal to fuse the image of one satellite with the nearest available image from another satellite. The best results are highlighted in \colorbox{\rcolor}{\textbf{green}} and the competing (second best) are represented using a \textbf{bold font}.


\begin{table*}[!ht]
\centering
 \begin{tabular}{l|c|c|c}
        \toprule
        \textbf{Satellite} & \textbf{L8} & \textbf{S2} & \textbf{S1} \\
        \midrule
        \textbf{OA Accuracy} & $69.69\% \pm 2.56\%$ & $\textbf{75.19}\% \pm \textbf{3.27}\%$ & \cellcolor{\rcolor}$\textbf{80.16}\% \pm \textbf{2.71}\%$ \\
        \textbf{Paddy Accuracy} & $54.85\% \pm 11.49\%$ & $62.45\% \pm 2.95\%$ & $72.18\% \pm 13.95\%$ \\
        \textbf{Non-Paddy Accuracy} & $74.53\% \pm 6.57\%$ & $80.29\% \pm 5.60\%$ & $84.23\% \pm 4.10\%$ \\
        \midrule
        \textbf{OA F1 Score} & $62.70\% \pm 1.89\%$ & $\textbf{70.64}\% \pm \textbf{2.60}\%$ & \cellcolor{\rcolor}$\textbf{77.78}\% \pm \textbf{3.86}\%$ \\
        \textbf{Paddy F1 Score} & $46.74\% \pm 4.58\%$ & $59.13\% \pm 2.29\%$ & $70.62\% \pm 6.45\%$ \\
        \textbf{Non-Paddy F1 Score} & $78.65\% \pm 2.81\%$ & $82.14\% \pm 2.95\%$ & $84.93\% \pm 1.51\%$ \\
        \midrule
        \textbf{OA IoU} & $47.73\% \pm 1.77\%$ & $\textbf{55.89}\% \pm \textbf{3.23}\%$ & \cellcolor{\rcolor}$\textbf{64.35}\% \pm \textbf{4.82}\%$ \\
        \textbf{Paddy IoU} & $30.59\% \pm 3.83\%$ & $42.01\% \pm 2.29\%$ & $54.87\% \pm 7.74\%$ \\
        \textbf{Non-Paddy IoU} & $64.88\% \pm 3.80\%$ & $69.77\% \pm 4.24\%$ & $73.83\% \pm 2.26\%$ \\
        \bottomrule
    \end{tabular}
\caption{\textbf{Single-Image Crop Segmentation task using U-Net 2D architecture.} All the images within the \textit{regional standard growing season} are used as separate input for the crop type segmentation task.}
\label{tab:unet2d-ct}
\end{table*}

\begin{table*}[]
\centering
\begin{tabular}{l|c|c|c}
\toprule
\textbf{Satellite} & {\textbf{L8}} & {\textbf{S2}} & {\textbf{S1}} \\
\midrule
\textbf{OA Accuracy} & $71.17\% \pm 1.64\%$ & $\textbf{74.67}\% \pm \textbf{2.37}\%$ & \cellcolor{\rcolor}$ \textbf{82.35}\% \pm \textbf{1.64}\%$ \\
\textbf{Paddy Accuracy} & $51.03\% \pm 11.93\%$ & $57.70\% \pm 10.15\%$ & $77.26\% \pm 8.44\%$ \\
\textbf{Non-Paddy Accuracy} & $77.76\% \pm 5.29\%$ & $81.47\% \pm 5.37\%$ & $84.94\% \pm 1.92\%$ \\
\midrule
\textbf{OA F1 Score} & $63.15\% \pm 2.58\%$ & $\textbf{69.21}\% \pm \textbf{2.58}\%$ & \cellcolor{\rcolor}$\textbf{80.51}\% \pm \textbf{2.39}\%$ \\
\textbf{Paddy F1 Score} & $46.09\% \pm 5.96\%$ & $56.35\% \pm 4.48\%$ & $74.55\% \pm 3.96\%$ \\
\textbf{Non-Paddy F1 Score} & $80.21\% \pm 1.89\%$ & $82.07\% \pm 2.11\%$ & $86.46\% \pm 0.84\%$ \\
\midrule
\textbf{OA IoU} & $48.53\% \pm 2.02\%$ & $\textbf{54.48}\% \pm \textbf{2.77}\%$ & \cellcolor{\rcolor}$\textbf{67.85}\% \pm \textbf{3.11}\%$ \\
\textbf{Paddy IoU} & $30.08\% \pm 4.85\%$ & $39.32\% \pm 4.38\%$ & $59.55\% \pm 4.95\%$ \\
\textbf{Non-Paddy IoU} & $66.98\% \pm 2.62\%$ & $69.64\% \pm 3.03\%$ & $76.15\% \pm 1.31\%$ \\
\bottomrule
\end{tabular}
\caption{\textbf{Single-Image Crop Segmentation task using DeepLabV3+ architecture.} All the images within the \textit{regional standard growing season} are used as separate input for the crop type segmentation task.}
\label{tab:deep-ct}
\end{table*}

\begin{table*}[!ht]
\centering
\begin{tabular}{l|c|c|c|c}
\toprule
\textbf{Satellite} & \textbf{L8} & \textbf{S2} & \textbf{S1} & \textbf{Fusion} \\
\midrule
OA Accuracy & 75.27\% $\pm$ 4.12\% & 89.06\% $\pm$ 1.80\% & \cellcolor{\rcolor}\textbf{91.05\%} $\pm$ \textbf{3.71\%} & \textbf{90.65\%} $\pm$ \textbf{3.18\%} \\
Paddy Accuracy & 54.99\% $\pm$ 14.66\% & 80.80\% $\pm$ 6.29\% & 82.46\% $\pm$ 8.41\% & 82.93\% $\pm$ 7.14\% \\
Non-Paddy Accuracy & 85.50\% $\pm$ 8.58\% & 93.23\% $\pm$ 1.32\% & 95.38\% $\pm$ 4.20\% & 94.54\% $\pm$ 3.39\% \\
\midrule
OA F1 Score & 70.57\% $\pm$ 5.40\% & 87.50\% $\pm$ 2.32\% & \cellcolor{\rcolor}\textbf{89.71\%} $\pm$ \textbf{4.20\%} & \textbf{89.32}\% $\pm$ \textbf{3.61\%} \\
Paddy F1 Score & 59.11\% $\pm$ 9.22\% & 83.10\% $\pm$ 3.46\% & 86.02\% $\pm$ 5.68\% & 85.56\% $\pm$ 4.88\% \\
Non-Paddy F1 Score & 82.02\% $\pm$ 3.56\% & 91.90\% $\pm$ 1.19\% & 93.40\% $\pm$ 2.78\% & 93.07\% $\pm$ 2.37\% \\
\midrule
OA IoU & 56.04\% $\pm$ 5.84\% & 78.12\% $\pm$ 3.48\% & \cellcolor{\rcolor}\textbf{81.77}\% $\pm$ \textbf{6.60}\% & \textbf{81.07}\% $\pm$ \textbf{5.77}\% \\
Paddy IoU & 42.43\% $\pm$ 8.99\% & 71.20\% $\pm$ 4.96\% & 75.81\% $\pm$ 8.51\% & 75.02\% $\pm$ 7.49\% \\
Non-Paddy IoU & 69.65\% $\pm$ 5.09\% & 85.04\% $\pm$ 2.04\% & 87.73\% $\pm$ 4.77\% & 87.12\% $\pm$ 4.13\% \\

\bottomrule
\end{tabular}
\caption{\textbf{Time-Series Crop Segmentation task using U-Net 3D architecture.} The time-series data prepared using the proposed method is used as an input for the crop type segmentation task.}
\label{tab:unet3d-ct}
\end{table*}

\begin{table*}[!ht]
\centering
\begin{tabular}{l|c|c|c|c}
\toprule
\textbf{Satellite} & \textbf{L8} & \textbf{S2} & \textbf{S1} & \textbf{Fusion} \\
\midrule
OA Accuracy & \cellcolor{\rcolor}$\textbf{77.03\%} \pm \textbf{4.00\%}$ & $70.87\% \pm 5.44\%$ & $75.33\% \pm 9.70\%$ & $\textbf{76.28\%} \pm \textbf{5.31\%}$ \\
Paddy Accuracy & $63.55\% \pm 28.73\%$ & $32.23\% \pm 15.44\%$ & $41.25\% \pm 31.55\%$ & $54.17\% \pm 5.24\%$ \\
Non-Paddy Accuracy & $83.83\% \pm 12.15\%$ & $90.34\% \pm 6.84\%$ & $92.52\% \pm 5.31\%$ & $87.42\% \pm 8.99\%$ \\
\midrule
OA F1 Score & \cellcolor{\rcolor}$\textbf{72.39\%} \pm \textbf{7.90\%}$ & $60.85\% \pm 8.43\%$ & $65.27\% \pm 18.52\%$ & $\textbf{71.81\%} \pm \textbf{4.64\%}$ \\
Paddy F1 Score & $61.98\% \pm 15.34\%$ & $41.24\% \pm 14.47\%$ & $46.96\% \pm 32.20\%$ & $60.75\% \pm 4.87\%$ \\
Non-Paddy F1 Score & $82.80\% \pm 3.28\%$ & $80.46\% \pm 3.74\%$ & $83.57\% \pm 5.56\%$ & $82.88\% \pm 4.72\%$ \\
\midrule
OA IoU & \cellcolor{\rcolor}$\textbf{58.52}\% \pm \textbf{8.38}\%$ & $47.15\% \pm 8.10\%$ & $53.64\% \pm 17.21\%$ & $\textbf{57.37}\% \pm \textbf{5.58}\%$ \\
Paddy IoU & $46.30\% \pm 15.70\%$ & $26.88\% \pm 12.41\%$ & $35.18\% \pm 26.91\%$ & $43.76\% \pm 4.93\%$ \\
Non-Paddy IoU & $70.75\% \pm 4.65\%$ & $67.43\% \pm 5.30\%$ & $72.09\% \pm 8.11\%$ & $70.97\% \pm 6.65\%$ \\

\bottomrule
\end{tabular}
\caption{\textbf{Time-Series Crop Segmentation task using ConvLSTM architecture.} The time-series data prepared using the proposed method is used as an input for the crop type segmentation task.}
\label{tab:conv-ct}
\end{table*}

\begin{table*}[!ht]
\centering
\begin{tabular}{l|c|c|c|c}
\toprule
\textbf{Satellite} & \textbf{L8} & \textbf{S2} & \textbf{S1} & \textbf{Fusion} \\
\midrule
OA Accuracy & 73.80\% $\pm$ 4.32\% & 66.36\% $\pm$ 0.20\% &\cellcolor{\rcolor} \textbf{82.12\%} $\pm$ \textbf{2.39\%} & \textbf{77.51\% $\pm$ 5.66\%} \\
Paddy Accuracy & 27.99\% $\pm$ 16.28\% & 0.04\% $\pm$ 0.09\% & 68.88\% $\pm$ 7.96\% & 54.68\% $\pm$ 19.18\% \\
Non-Paddy Accuracy & 96.90\% $\pm$ 3.16\% & 99.80\% $\pm$ 0.31\% & 88.80\% $\pm$ 2.09\% & 89.02\% $\pm$ 12.47\% \\
\midrule
OA F1 Score & 61.13\% $\pm$ 12.06\% & 39.93\% $\pm$ 0.12\% & \cellcolor{\rcolor}\textbf{79.39\%} $\pm$ \textbf{3.09\%} & \textbf{72.23\% $\pm$ 7.03\%} \\
Paddy F1 Score & 39.09\% $\pm$ 22.23\% & 0.08\% $\pm$ 0.18\% & 71.92\% $\pm$ 4.69\% & 60.70\% $\pm$ 11.72\% \\
Non-Paddy F1 Score & 83.16\% $\pm$ 2.10\% & 79.78\% $\pm$ 0.15\% & 86.87\% $\pm$ 1.61\% & 83.77\% $\pm$ 5.39\% \\
\midrule
OA IoU & 48.58\% $\pm$ 8.83\% & 33.20\% $\pm$ 0.11\% & \cellcolor{\rcolor}\textbf{66.57}\% $\pm$ \textbf{4.17}\% & \textbf{58.37}\% $\pm$ \textbf{8.06}\% \\
Paddy IoU & 25.94\% $\pm$ 14.92\% & 0.04\% $\pm$ 0.09\% & 56.33\% $\pm$ 5.96\% & 44.38\% $\pm$ 12.06\% \\
Non-Paddy IoU & 71.21\% $\pm$ 3.05\% & 66.36\% $\pm$ 0.20\% & 76.81\% $\pm$ 2.56\% & 72.35\% $\pm$ 7.60\% \\

\bottomrule
\end{tabular}
\caption{\textbf{Time-Series Crop Segmentation task using U-TAE architecture.} The time-series data prepared using the proposed method is used as an input for the crop type segmentation task.}
\label{tab:utae-ct}
\end{table*}

\begin{table*}[!ht]
\centering
\begin{tabular}{lcccc}
\toprule
\textbf{Satellite} & \textbf{RMSE} & \textbf{MAE} & \textbf{MAPE} \\
\midrule
\textbf{L8} & $3.63 \pm 0.64$ & $2.66 \pm 0.96$ & $1.45\% \pm 0.53\%$ \\
\textbf{S2} & $3.22 \pm 0.59$ & \cellcolor{\rcolor}$\textbf{2.30} \pm \textbf{0.61}$ & $1.26\% \pm 0.33\%$ \\
\textbf{S1} & $4.50 \pm 1.14$ & $3.61 \pm 0.90$ & $1.97\% \pm 0.49\%$ \\
\textbf{Fusion} & $3.61 \pm 0.59$ & $\textbf{2.33} \pm \textbf{0.64}$ & $1.27\% \pm 0.35\%$ \\
\bottomrule
\end{tabular}
\caption{\textbf{Sowing Date prediction task using U-Net 3D architecture. }The time-series data prepared using the proposed method is used as an input for predicting the sowing date of the paddy crops.}
\label{tab:unet3d-sd}
\end{table*}

\begin{table*}[!ht]
\centering
 \begin{tabular}{lccc}
        \toprule
        \textbf{Satellite}& \textbf{RMSE} & \textbf{MAE} & \textbf{MAPE} \\
        \hline
        \textbf{L8} & $6.23 \pm 1.839$ & $5.13 \pm 1.821$ & $2.81\% \pm 1.00\%$ \\
        \textbf{S2} & $3.38 \pm 0.075$ & \cellcolor{\rcolor}$\textbf{2.90} \pm \textbf{0.111}$ & $1.59\% \pm 0.06\%$ \\
        \textbf{S1} & $4.94 \pm 0.231$ & $3.77 \pm 0.519$ & $2.06\% \pm 0.28\%$ \\
        \textbf{Fusion} & $3.61 \pm 0.687$ & $\textbf{2.91} \pm \textbf{0.454}$ & $1.59\% \pm 0.25\%$ \\
        \bottomrule
    \end{tabular}
\caption{\textbf{Sowing Date prediction task using ConvLSTM architecture.} The time-series data prepared using the proposed method is used as an input for predicting the sowing date of the paddy crops.}
\label{tab:conv-sd}
\end{table*}

\begin{table*}[!ht]
\centering
\begin{tabular}{lccc}
\toprule
\textbf{Satellite} & \textbf{RMSE} & \textbf{MAE} & \textbf{MAPE} \\
\hline
\textbf{L8} & $5.72 \pm 0.36$ & $3.88 \pm 0.33$ & $2.12\% \pm 0.18\%$ \\
\textbf{S2 } & $3.55 \pm 0.36$ & $\textbf{2.97} \pm \textbf{0.34}$ & $1.62\% \pm 0.19\%$ \\
\textbf{S1 } & $4.95 \pm 0.10$ & $3.22 \pm 0.15$ & $1.76\% \pm 0.08\%$ \\
\textbf{Fusion} & $3.86 \pm 0.71$ & \cellcolor{\rcolor}$\textbf{2.91} \pm \textbf{0.62}$ & $1.59\% \pm 0.34\%$ \\
\bottomrule
\end{tabular}
\caption{\textbf{Sowing Date prediction task using U-TAE architecture.} The time-series data prepared using the proposed method is used as an input for predicting the sowing date of the paddy crops.}
\label{tab:utae-sd}
\end{table*}

\begin{table*}[!ht]
\centering
\begin{tabular}{lccc}
\toprule
\textbf{Satellite} & \textbf{RMSE } & \textbf{MAE} & \textbf{MAPE } \\
\hline
\textbf{L8} & 9.50 $\pm$ 2.635 & \textbf{6.20} $\pm$ \textbf{1.030} & 3.39\% $\pm$ 0.56\% \\
\textbf{S2} & 9.45 $\pm$ 2.475 & 6.36 $\pm$ 2.164 & 3.48\% $\pm$ 1.18\% \\
\textbf{S1} & 10.67 $\pm$ 1.039 & 7.23 $\pm$ 0.779 & 3.95\% $\pm$ 0.43\% \\
\textbf{Fusion} & 9.28 $\pm$ 2.312 & \cellcolor{\rcolor}\textbf{6.16} $\pm$ \textbf{1.770} & 3.37\% $\pm$ 0.97\% \\
\bottomrule
\end{tabular}
\caption{\textbf{Transplanting Date prediction task using U-Net 3D architecture.} The time-series data prepared using the proposed method is used as an input for predicting the transplanting date of the paddy crops.}
\label{tab:unet3d-td}
\end{table*}

\begin{table*}[!ht]
\centering
\begin{tabular}{lccc}
\toprule
\textbf{Satellite} & \textbf{RMSE} & \textbf{MAE } & \textbf{MAPE } \\
\hline
\textbf{L8} & 14.50 $\pm$ 0.93 & 9.41 $\pm$ 0.48 & 5.14\% $\pm$ 0.26\% \\
\textbf{S2} & 11.17 $\pm$ 0.31 & \textbf{7.47} $\pm$ \textbf{0.27} & 4.08\% $\pm$ 0.15\% \\
\textbf{S1} & 12.11 $\pm$ 0.90 & 8.37 $\pm$ 0.53 & 4.57\% $\pm$ 0.29\% \\
\textbf{Fusion} & 11.62 $\pm$ 0.93 & \cellcolor{\rcolor}\textbf{7.34} $\pm$ \textbf{0.59} & 4.01\% $\pm$ 0.32\% \\
\bottomrule
\end{tabular}
\caption{\textbf{Transplanting Date prediction task using ConvLSTM architecture.} The time-series data prepared using the proposed method is used as an input for predicting the transplanting date of the paddy crops.}
\label{tab:conv-td}
\end{table*}

\begin{table*}[!ht]
\centering
\begin{tabular}{lccc}
\toprule
\textbf{Satellite} & \textbf{RMSE}  & \textbf{MAE}   & \textbf{MAPE}   \\
\hline
\textbf{L8 } & $11.56 \pm 1.25$ & $8.75 \pm 0.97$ & $4.78\% \pm 0.53\%$ \\
\textbf{S2 } & $10.66 \pm 0.78$ & $\textbf{7.33} \pm \textbf{0.76}$ & $4.01\% \pm 0.41\%$ \\
\textbf{S1} & $11.38 \pm 0.14$ & $7.60 \pm 0.18$ & $4.15\% \pm 0.10\%$ \\
\textbf{Fusion } & $9.83 \pm 1.69$ & \cellcolor{\rcolor}$\textbf{6.62} \pm \textbf{1.23}$ & $3.62\% \pm 0.67\%$ \\
\bottomrule
\end{tabular}
\caption{\textbf{Transplanting Date prediction task using U-TAE architecture.} The time-series data prepared using the proposed method is used as an input for predicting the transplanting date of the paddy crops.}
\label{tab:utae-td}
\end{table*}

\begin{table*}[!ht]
\centering
\begin{tabular}{lcccc}
\toprule
\textbf{Satellite} & \textbf{RMSE } & \textbf{MAE } & \textbf{MAPE} \\
\midrule
\textbf{L8}       & $12.99 \pm 1.77$ & $\textbf{9.86} \pm \textbf{0.74}$ & $5.39\% \pm 0.40\%$ \\
\textbf{S2}       & $11.51 \pm 1.69$ & \cellcolor{\rcolor}$\textbf{8.83} \pm \textbf{1.52}$ & $4.82\% \pm 0.83\%$ \\
\textbf{S1}       & $13.17 \pm 0.67$ & $10.08 \pm 0.56$ & $5.51\% \pm 0.31\%$ \\
\textbf{Fusion}   & $14.09 \pm 3.88$ & $10.75 \pm 3.39$ & $5.87\% \pm 1.85\%$ \\
\bottomrule
\end{tabular}
\caption{\textbf{Harvesting Date prediction task using U-Net 3D architecture.} The time-series data prepared using the proposed method is used as an input for predicting the harvesting date of the paddy crops.}
\label{tab:unet3d-hd}
\end{table*}

\begin{table*}[!ht]
\centering
\begin{tabular}{lccc}
\toprule
\textbf{Satellite} & \textbf{RMSE} & \textbf{MAE} & \textbf{MAPE} \\
\midrule
\textbf{L8} & $23.97 \pm 2.71$ & $20.14 \pm 2.29$ & $11.00\% \pm 1.25\%$ \\
\textbf{S2} & $19.30 \pm 4.26$ & $\textbf{16.40} \pm \textbf{3.32}$ & $8.96\% \pm 1.82\%$ \\
\textbf{S1} & $19.92 \pm 2.71$ & $17.52 \pm 2.10$ & $9.57\% \pm 1.14\%$ \\
\textbf{Fusion} & $17.92 \pm 1.82$ & \cellcolor{\rcolor}$\textbf{15.00} \pm \textbf{1.72}$ & $8.20\% \pm 0.94\%$ \\
\bottomrule
\end{tabular}
\caption{\textbf{Harvesting Date prediction task using ConvLSTM architecture.} The time-series data prepared using the proposed method is used as an input for predicting the harvesting date of the paddy crops.}
\label{tab:conv-hd}
\end{table*}

\begin{table*}[!ht]
\centering
\begin{tabular}{lccc}
\toprule
\textbf{Satellite} & \textbf{RMSE} & \textbf{MAE} & \textbf{MAPE} \\
\midrule
\textbf{L8} & $18.61 \pm 2.49$ & $15.28 \pm 2.11$ & $8.35\% \pm 1.15\%$ \\ 
\textbf{S2} & $13.47 \pm 1.52$ & $11.10 \pm 1.13$ & $6.06\% \pm 0.62\%$ \\ 
\textbf{S1} & $13.01 \pm 0.40$ & \cellcolor{\rcolor}$\textbf{9.97} \pm \textbf{0.54}$ & $5.45\% \pm 0.30\%$ \\ 
\textbf{Fusion} & $13.10 \pm 1.13$ & $\textbf{10.48} \pm \textbf{1.24}$ & $5.73\% \pm 0.67\%$ \\ 
\bottomrule
\end{tabular}
\caption{\textbf{Harvesting Date prediction task using U-TAE architecture.} The time-series data prepared using the proposed method is used as an input for predicting the harvesting date of the paddy crops.}
\label{tab:utae-hd}
\end{table*}

\begin{table*}[!ht]
\centering
 \begin{tabular}{lccc}
        \toprule
        \textbf{Satellite} & \textbf{RMSE } & \textbf{MAE } & \textbf{MAPE} \\
        \midrule
        \textbf{L8} & $1165.91 \pm 112.10$ & $860.61 \pm 71.31$ & \cellcolor{\rcolor}$\textbf{46.74}\% \pm \textbf{3.82}\%$ \\
        \textbf{S2} & $1462.35\pm 242.51$ & $1080.82 \pm 245.65$ & $60.44\% \pm 14.50\%$ \\
        \textbf{S1} & $1206.86\pm 121.01$ & $877.50 \pm 121.64$ & $\textbf{48.35}\% \pm \textbf{7.64}\%$ \\
        \bottomrule
    \end{tabular}

\caption{\textbf{Single image crop yield prediction task using U-Net 2D architecture.} Image that is available just before the harvesting date is used for estimating the yield of the paddy crops.}
\label{tab:unet2d-cy}
\end{table*}

\begin{table*}[!ht]
\centering
\begin{tabular}{lccc}
\toprule
\textbf{Satellite} & \textbf{RMSE} & \textbf{MAE} & \textbf{MAPE} \\
\hline

\textbf{L8} & $1222.90 \pm 59.27$  & $877.65 \pm 36.90$   & \cellcolor{\rcolor}$\textbf{46.90}\% \pm \textbf{1.80}\%$ \\
\textbf{S2} & $1526.32 \pm 240.54$ & $1099.90 \pm 160.51$ & $63.19\% \pm 11.01\%$ \\
\textbf{S1} & $1412.41 \pm 137.34$ & $1016.65 \pm 81.09$  & $\textbf{55.37}\% \pm \textbf{4.08}\%$ \\
\bottomrule
\end{tabular}

\caption{\textbf{Single image crop yield prediction task using DeepLabV3+ architecture.} Image that is available just before the harvesting date is used as separate input for estimating the yield of the paddy crops.}
\label{tab:deep-cy}
\end{table*}

\begin{table*}[!ht]
\centering
\begin{tabular}{lccc}
\hline
\textbf{Satellite}  & \textbf{RMSE} & \textbf{MAE } & \textbf{MAPE} \\
\hline
\textbf{L8} & $1353.83 \pm 151.97$ & $968.43 \pm 129.89$ & \cellcolor{\rcolor}$\textbf{54.00}\% \pm \textbf{9.67}\%$ \\
 \textbf{S2} & $1574.40 \pm 113.52$ & $1209.35 \pm 87.00$ & $72.38\% \pm 8.74\%$ \\
 \textbf{S1} & $1533.31 \pm 217.31$ & $1144.96 \pm 195.39$ & $71.81\% \pm 17.27\%$ \\
 \textbf{Fusion} & $1484.82 \pm 239.34$ & $1131.55 \pm 194.96$ & $\textbf{70.35}\% \pm \textbf{13.75}\%$ \\
\bottomrule
\end{tabular}

\caption{\textbf{Crop yield prediction with regional standard season using U-Net 3D architecture.} The time-series data prepared using the proposed method is used as an input for estimating the yield of the paddy crops when using the \textit{regional standard growing season}.}
\label{tab:unet3d-cy}
\end{table*}

\begin{table*}[!ht]
\centering
\begin{tabular}{lcccc}
\hline
        \textbf{Satellite} & \textbf{RMSE } & \textbf{MAE } & \textbf{MAPE } \\
        \hline
        \textbf{L8} & $88.37 \pm 4.70\%$ & $70.79 \pm 6.80\%$ & $64.10\% \pm 4.70\% $ \\
        \textbf{S2} & $74.34 \pm 4.70\%$ & $52.27 \pm 6.80\%$ & $\textbf{62.75}\% \pm \textbf{6.80}\% \pm $ \\
        \textbf{S1} & $41.70 \pm 4.70\%$ & $33.02 \pm 6.80\%$ &  \cellcolor{\rcolor}$\textbf{59.98}\% \pm \textbf{5.34}\% $ \\
        \textbf{Fusion} & $72.27 \pm 4.70\%$ & $31.01 \pm 6.80\%$ & $63.89\% \pm 3.20\%  $ \\
        \hline
    \end{tabular}
\caption{\textbf{Crop yield prediction with regional standard season using ConvLSTM architecture.} The time-series data prepared using the proposed method is used as an input for estimating the yield of the paddy crops when using the \textit{regional standard growing season}.}
\label{tab:conv-cy}
\end{table*}

\begin{table*}[!ht]
\centering
\begin{tabular}{lccc}
\hline

\textbf{Satellite} & \textbf{RMSE} & \textbf{MAE} & \textbf{MAPE} \\
\hline
\textbf{L8} & $1294.97 \pm 85.22$ & $908.09 \pm 54.81$ & $\textbf{50.40}\% \pm \textbf{4.41}\%$ \\
\textbf{S2} & $1342.07 \pm 66.27$ & $985.42 \pm 69.44$ & $56.51\% \pm 6.71\%$ \\
\textbf{S1} & $1331.32 \pm 90.26$ & $914.01 \pm 61.61$ & $52.78\% \pm 8.12\%$ \\
\textbf{Fusion} & $1242.00 \pm 119.18$ & $884.60 \pm 103.12$ &  \cellcolor{\rcolor}$\textbf{49.63}\% \pm \textbf{7.95}\%$ \\
\bottomrule
\end{tabular}
\caption{\textbf{Crop yield prediction with regional standard season using U-TAE architecture.} The time-series data prepared using the proposed method is used as an input for estimating the yield of the paddy crops when using the \textit{regional standard growing season}.}
\label{tab:utae-cy}
\end{table*}

\begin{table*}[!ht]
\centering
\begin{tabular}{lccc}
\hline

\textbf{Satellite} & \textbf{RMSE} & \textbf{MAE} & \textbf{MAPE} \\
\hline
\textbf{L8} & $1402.42 \pm 306.53$ & $1024.57 \pm 215.73$ & \cellcolor{\rcolor}$\textbf{59.38}\% \pm \textbf{14.75}\%$ \\
\textbf{S2} & $1590.68 \pm 155.93$ & $1211.72 \pm 139.84$ & $73.59\% \pm 9.81\%$ \\
\textbf{S1} & $1493.99 \pm 257.70$ & $1096.79 \pm 223.64$ & $65.66\% \pm 16.24\%$ \\
\textbf{Fusion} & $1426.34 \pm 187.12$ & $1085.44 \pm 202.35$ & $\textbf{64.56}\% \pm \textbf{13.77}\%$ \\
\hline

\end{tabular}
\caption{\textbf{Crop yield prediction with actual growing season using U-Net 3D architecture.} The time-series data prepared using the proposed method is used as an input for estimating the yield of the paddy crops when using the \textit{actual standard growing season}.}
\label{tab:unet3d-cy-as}
\end{table*}

\begin{table*}[!ht]
\centering
\begin{tabular}{lccc}
\hline

\textbf{Satellite}  & \textbf{RMSE } & \textbf{MAE } & \textbf{MAPE } \\
\hline
\textbf{L8} & $1353.81 \pm 142.08$ & $990.63 \pm 113.83$ & $59.92\% \pm 10.60\%$ \\
\textbf{S2} & $1411.31 \pm 105.86$ & $1019.06 \pm 76.15$ & $60.83\% \pm 6.71\%$ \\
\textbf{S1} & $1260.50 \pm 62.30$ & $896.78 \pm 62.11$ & \cellcolor{\rcolor}$\textbf{52.46}\% \pm \textbf{6.36}\%$ \\
\textbf{Fusion}& $1540.05 \pm 276.92$ & $1146.27 \pm 203.53$ & $72.35\% \pm 12.67\%$ \\
\hline

\end{tabular}
\caption{\textbf{Crop yield prediction with actual growing season using ConvLSTM architecture.} The time-series data prepared using the proposed method is used as an input for estimating the yield of the paddy crops when using the \textit{actual standard growing season}.}
\label{tab:conv-cy-as}
\end{table*}

\begin{table*}[!ht]
\centering
\begin{tabular}{lccc}
\hline
\textbf{Satellite} & \textbf{RMSE} & \textbf{MAE} & \textbf{MAPE} \\
\hline
\textbf{L8} & $1370.25 \pm 109.90$ & $967.62 \pm 86.067$ & \cellcolor{\rcolor}$\textbf{56.34}\% \pm \textbf{7.95}\%$ \\
\textbf{S2} & $1374.01 \pm 167.84$ & $1008.11 \pm 162.92$ & $58.11\% \pm 12.19\%$ \\
\textbf{S1} & $1478.81 \pm 148.82$ & $1019.66 \pm 136.36$ & $61.30\% \pm 10.13\%$ \\
\textbf{Fusion} & $1327.31 \pm 213.83$ & $961.15 \pm 139.91$ & \cellcolor{\rcolor}$\textbf{57.61}\% \pm \textbf{14.10}\%$ \\
\hline
\end{tabular}
\caption{\textbf{Crop yield prediction with actual growing season using U-TAE architecture.} The time-series data prepared using the proposed method is used as an input for estimating the yield of the paddy crops when using the \textit{actual standard growing season}.}
\label{tab:utae-cy-as}
\end{table*}

{\small
\bibliographystyle{ieee_fullname}
\bibliography{egbib}
}